\documentclass[11pt]{article}

\usepackage[preprint]{acl}

\usepackage{times}
\usepackage{latexsym}

\usepackage[T1]{fontenc}
\usepackage[utf8]{inputenc}

\usepackage{microtype}
\usepackage{inconsolata}

\usepackage{graphicx}

\usepackage{amsmath}
\usepackage{amsfonts}
\usepackage{booktabs}
\usepackage{multirow}

\usepackage[most]{tcolorbox}
\usepackage{xcolor} 
\usepackage{tabularx}
\usepackage{booktabs}
\usepackage{fontawesome5}
\usepackage{amsthm}

\newtheorem*{definition*}{Definition}

\newtcolorbox{logicblock}{ enhanced, breakable, colback=gray!3, colframe=gray!35, boxrule=0.5pt, arc=1pt, left=5pt, right=5pt, top=5pt, bottom=5pt, before upper={\small} }

\title{Logic Before Language: Pre-pretraining on Formal Derivations Fosters Skill Acquisition and Compressibility}

\author{
\textbf{Jo-Ku Cheng}
\quad
\textbf{Nikolaos Aletras}
\quad
\textbf{Marco Valentino}
\\
School of Computer Science, University of Sheffield, United Kingdom
\\
\texttt{\{jcheng34,n.aletras,m.valentino\}@sheffield.ac.uk}
}

\begin{document}
\maketitle
\begin{abstract}

Pre-pretraining language models (LMs) on symbolic data can accelerate and improve natural language acquisition. However, existing pre-pretraining tasks, such as Dyck and procedural algorithms, rely on narrow primitives that fail to capture the expressive capacity of natural language. Moreover, prior studies remain restricted to relatively small token budgets, offering limited insight into skill emergence and representational dynamics. To address these limitations, we propose logic pre-pretraining (Logic-PPT) as a principled initialization strategy, leveraging formal derivations to impart richer structural and linguistic biases. Formal derivations require abstract mechanisms that are central to natural language, simultaneously binding variables, connecting quantifiers and relational dependencies, and composing predicate-argument structures over long contexts. Scaling our evaluation to a 100B-token regime, logic pre-pretraining substantially accelerates skill acquisition in LMs, achieving 80\% accuracy on linguistic tasks with 36B fewer tokens than standard initialization, and outperforming alternative pre-pretraining baselines. Mechanistically, formal derivations induce persistent structural reorganization, distinctively characterized by a lower-rank, spectrally concentrated representation space. Crucially, we show that this internal geometry enables improved model compressibility via pruning, matching the dense baseline performance even at $\approx$33\% sparsity.  
\end{abstract}

\section{Introduction}

\begin{figure}
    \centering
    \includegraphics[width=\linewidth]{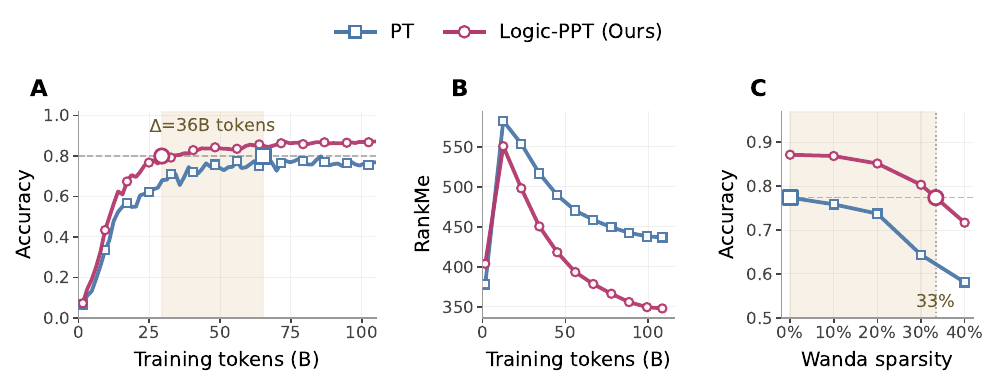}
    \caption{
    \textbf{(A) Faster learning:} Logic-PPT reaches 0.8 mean accuracy using 36B fewer natural language training tokens than the PT baseline in the 100B-token run.
    \textbf{(B) Lower-rank representations:} Logic-PPT promotes a more structured and compact internal geometry, organizing representations around a smaller set of directions.
    \textbf{(C) Greater compression robustness:} At $\approx$33\% sparsity, the Logic-PPT model retains accuracy comparable to a dense PT baseline.
    }
    \label{fig:overview}
\end{figure}

Recent work on \textit{symbolic pre-pretraining}, exposing language models (LMs) to procedurally generated symbolic data prior to natural language training, has emerged as a  strategy to accelerate natural language acquisition and improve downstream performance and robustness to noise~\cite{hu-etal-2025-circuits,jiang2026procedural, guo2026syntheticprepretrainingimproveslanguage, lee2026traininglanguagemodelsneural,mita-etal-2026-language}.

However, existing pre-pretraining frameworks remain limited in two critical respects. Qualitatively, the symbolic tasks considered in prior work capture only a narrow range of structural inductive biases relevant to language\cite{wilcox-etal-2018-rnn, doi:https://doi.org/10.1002/9780470758335.ch1, hewitt-manning-2019-structural}. For example, Dyck languages primarily isolate hierarchical nesting and bracket symmetry~\cite{hu-etal-2025-circuits}, whereas algorithmic tasks such as sorting isolate purely procedural sequence transformations, lacking the expressive capacity inherent to natural language~\cite{jiang2026procedural}. Quantitatively, prior studies have generally trained on fewer than 10B natural language tokens~\cite{hu-etal-2025-circuits, lee2026traininglanguagemodelsneural, jiang2026procedural}, leaving unclear whether the impact of symbolic pre-pretraining  persists at longer training and whether it fundamentally changes internal representational dynamics.

These limitations motivate our contribution: \textit{we introduce formal logical derivations as a more expressive and structurally richer source of pre-pretraining data and study its impact on imparting linguistic biases and representational changes at larger training token budget (i.e. 100B tokens)}.

\begin{figure*}[t]
    \includegraphics[width=1\linewidth]{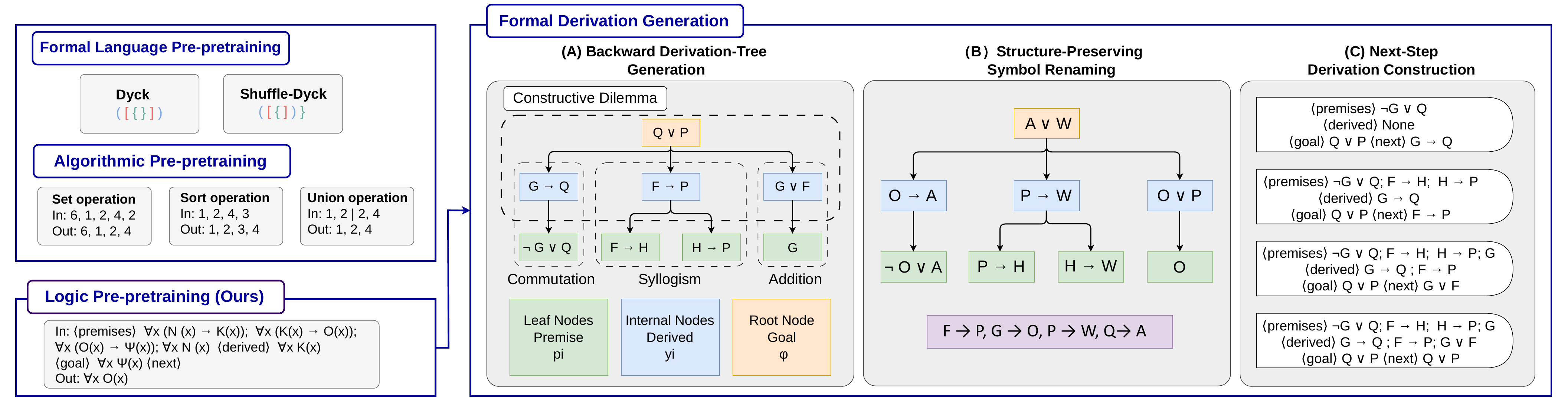}
    \caption{(Left) We introduce \textit{Logic Pre-pretraining (Logic-PPT)}, a new framework for initializing LMs on symbolic data before exposure to natural language. Compared to previous approaches relying on formal languages and procedural algorithms, Logic-PPT adopts more expressive and structurally richer formal derivations to impart deeper linguistic inductive biases.
    (Right) Overview of the methodology to generate synthetic formal derivations for Logic-PPT:
    (A) A derivation-tree is constructed by backward chain of rule schemata, starting from a goal and recursively replacing selected leaves with instantiated premises.
    (B) Each derivation-tree is augmented with symbol-renaming variants that preserve its topology and rule sequence.
    (C) A post-order traversal produces a leaf-to-root derivation sequence to construct a next-step prediction example.
      }
    \label{fig:pipeline}
\end{figure*}

Specifically, we organize our investigation around three questions: (1) \textit{Behavioral}: Can pre-pretraining on formal derivations accelerate and improve language and skill acquisition in LMs? (2) \textit{Mechanistic}: How does pre-pretraining affect internal representational geometry and dynamics? (3) \textit{Interventional}: can we leverage the structural changes induced by pre-pretraining to further improve LMs' efficiency?

To answer these questions, we curate a library of 247 formal logic schemata spanning propositional, term, and first-order logic. By leveraging backward chaining algorithms, we construct a large-scale next-step formal derivation dataset to symbolically pre-pretrain Transformer models prior to a 100B-token natural language pretraining run on FineWeb-Edu \cite{lozhkov2024fineweb-edu}. We systematically compare our logic pre-pretraining framework (\textit{Logic-PPT}) against established symbolic methods, including formal languages (Dyck/ShuffleDyck) and procedural algorithms (Set/Sort/Union), as well as a random initialization baseline. Our key findings can be summarized as follows:

\begin{itemize}
\item Logic-PPT significantly accelerates language and skill acquisition, achieving 80\% accuracy across a suite of linguistic tasks using 36B fewer natural language tokens than standard initialization and outperforming existing symbolic PPT methods by 7.1\% points in accuracy at the final checkpoint.

\item Mechanistically, formal derivations induce a distinctive lower-rank, spectrally concentrated representation space characterized by clearer layer differentiation that persists throughout 100B tokens of language pretraining.
\item We demonstrate for the first time that this internal geometry translates into superior pruning resilience, matching dense baseline performance at $\approx 33\%$ sparsity. At 40\% sparsity, Logic-PPT incurs only a 14.4\% relative drop compared to an average 26.8\% degradation across alternative symbolic PPT methods.
\end{itemize}

\section{Methodology}

Our objective is to investigate how pre-pretraining on formal logical derivations shapes subsequent natural language acquisition. Specifically, we examine whether formal logic instills inductive biases that accelerate skill emergence, reorganize internal representations, and enhance model efficiency. Our framework consists of two main stages: \textit{Logic Pre-Pretraining (Logic-PPT)} and \textit{Natural Language Pretraining (PT)} via backbone transfer.  

\subsection{Defining Formal Derivations}
Formal logical derivations demonstrate the validity of an argument by deriving conclusions from premises through a sequence of rule-governed steps~\cite{hardegree1994symbolic}. Each step is either one of the original premises or a formula that follows from preceding steps according to a derivation rule, with the final step being the conclusion.

\begin{definition*}

Let $\Gamma$ be a set of premises, $\mathcal{A}$ a set of logical axioms, and $\mathcal{R}$ a set of inference rules. A formal logical derivation of a formula $\varphi$ from $\Gamma$ is a finite sequence of formulas \[ D = \langle \alpha_0, \ldots, \alpha_n \rangle \] such that $\alpha_n$ is $\varphi$ and, for each $k \leq n$, either 

\begin{enumerate} 
\item $\alpha_k \in \Gamma \cup \mathcal{A}$, or 
\item $\alpha_k$ is obtained from formulas occurring earlier in the sequence by applying an inference rule in $\mathcal{R}$. 
\end{enumerate} 

If such a derivation exists, we write $\Gamma \vdash_{\mathcal{R}} \varphi$. 
\end{definition*}

\subsection{Stage 1: Logic Pre-pretraining}
\label{sec:logic_ppt}
In the first stage, the model is trained exclusively on formal derivations. We denote the randomly initialized model parameters as
$\Theta_{\text{logic}} = \{\Theta_{\text{backbone}}, \mathbf{E}_{\text{logic}}, \mathbf{H}_{\text{logic}}\}$, where $\Theta_{\text{backbone}}$ represents the shared Transformer backbone parameters (multi-head self-attention layers, feed-forward networks, and layer normalizations), $\mathbf{E}_{\text{logic}} \in \mathbb{R}^{\vert{}V_{\text{logic}}\vert{} \times d}$ is the input token embedding matrix, and $\mathbf{H}_{\text{logic}}\in \mathbb{R}^{d \times \vert{}V_{\text{logic}}\vert{}}$ is the output language-modeling head. Here, $d$ denotes the hidden dimension and $V_{\text{logic}}$ represents a character-level logic vocabulary.

\paragraph{Task Formulation.}
We cast formal derivation generation as an auto-regressive next-step prediction task. Given a derivation
$D=\langle \alpha_1,\ldots,\alpha_n\rangle$,
we construct one training example for each step $i$, where the input contains the premises $\Gamma=\{\ p_1,\ldots,p_m \}\ $, the derivation prefix $D_{<i} = \langle \alpha_1,\ldots,\alpha_{i-1}\rangle $, and the target conclusion $\varphi$, and its output is the next derivation step $y_i=\alpha_i$. The input is serialized as the token sequence $x_i$:

{
\small
\begin{equation}
  \begin{aligned}
  x_i ={}&
  \langle\mathtt{premises}\rangle\;
  p_1;\ldots;p_m                                      \\
  &\langle\mathtt{derived}\rangle\;
  \alpha_1;\ldots;\alpha_{i-1}                                  \\
  &\langle\mathtt{goal}\rangle\; \varphi\;                 \\
  &\langle\mathtt {next}\rangle \; .
  \end{aligned}
\end{equation}
}

This construction yields a training pair $(x_i, y_i)$. All sequences are processed using a domain-specific character-level tokenizer.  

\paragraph{Optimization Objective.} The model is trained autoregressively to predict the target step $y_i$. To prevent the model from expending capacity on reconstructing fixed prompt contexts, the cross-entropy loss is computed strictly over the tokens corresponding to the target step $y_i$:

{
\small
\begin{equation}
    \mathcal{L}_{\text{logic}}(\Theta_{\text{logic}}) = -\sum_{j=1}^{|y_i|} \log P_{\Theta_0}\big(y_{i, j} \mid x_i, y_{i, <j}\big).
\end{equation} 
}

After Logic-PPT, the resulting parameters are denoted as
$ 
\Theta_{\text{logic}}^{*}
=
 \{\
\Theta_{\text{backbone}}^{\text{logic}},
\mathbf{E}_{\text{logic}}^{*},
\mathbf{H}_{\text{logic}}^{*}
\}\ $.

\subsection{Stage 2: Natural Language Pretraining}
Following Logic-PPT, we transfer the inductive biases learned in the Transformer backbone to a standard natural language pretraining setup.  

\paragraph{Vocabulary \& Head Re-initialization.} Because symbolic logic and natural language operate over distinct token spaces, vocabulary-dependent parameters are non-transferable. We discard the symbolic token embedding matrix $\mathbf{E}_{\text{logic}}^{*}$ and output head $\mathbf{H}_{\text{logic}}^{*}$. We retain the pre-trained Transformer backbone parameters $\Theta_{\text{backbone}}^{\text{logic}}$ and attach a randomly initialized embedding matrix $\mathbf{E}_{\text{lang}} \in \mathbb{R}^{\vert{}V_{\text{lang}}\vert{} \times d}$ and output head $\mathbf{H}_{\text{lang}} \in \mathbb{R}^{d \times \vert{}V_{\text{lang}}\vert{}}$, sized to match a natural language tokenizer vocabulary $V_{\text{lang}}$.

\paragraph{Natural Language Pretraining.} The composite parameter set $\Theta = \{\ \Theta_{\text{backbone}}^{\text{logic}}, \mathbf{E}_{\text{lang}}, \mathbf{H}_{\text{lang}}\}$ is trained on a large-scale natural language text corpus $\mathcal{D}_{\text{text}}$ (FineWeb-Edu). All parameters remain fully trainable under the standard causal language modeling objective:

{
\small
\begin{equation}
\mathcal{L}_{\text{lang}}(\Theta) = -\sum_{t=1}^{T} \log P_{\Theta}\big(w_t \mid w_{<t}\big),
\end{equation}
}

where $w_t \in V_{\text{lang}}$ denotes the $t$-th natural language token in a sequence of length $T$.
 
\subsection{Logic Pre-pretraining Data}

\begin{table}[t]
    \centering
    \scriptsize
    \begin{tabular}{p{2.4cm}ll}
      \toprule\textbf{Schema}
      & \textbf{Premises}
      & \textbf{Conclusion} \\
      \midrule

      Propositional\\
      \midrule
      Modus ponens
      & \(P\rightarrow Q,\;P\)
      & \(Q\) \\
       Modus tollens
      & \(P\rightarrow Q,\;\neg Q\)
      & \(\neg P\) \\
      Resolution
      & \(P\lor Q,\;\neg P\lor R\)
      & \(Q\lor R\) \\

      \midrule

      Term\\
      \midrule
      Barbara
      & \(A\models B,\;B\models C\)
      & \(A\models C\) \\ Ferio
      & \(C\models\neg B,\;A\not\models\neg C\)
      & \(A\not\models B\) \\
      CM-Barbara
      & \(A\models\neg B,\;\neg B\models C\)
      & \(A\models C\) \\

      \midrule
        
      First-order\\
      \midrule
       Universal instantiation
      & \(\forall x\,\Phi(x)\)
      & \(\Phi(t)\) \\
       Universal modus ponens
      & \(\forall x(\Phi(x)\rightarrow\Psi(x)),\;\Phi(a)\)
      & \(\Psi(a)\) \\
      Equality substitution
      & \(s=t,\;\Phi(s)\)
      & \(\Phi(t)\) \\    
      \bottomrule
    \end{tabular}
\caption{Representative formal derivation schemata from the three logic families used for Logic-PPT.}
    \label{tab:scheme-sample}
  \end{table}

\paragraph{Collecting Formal  Schemata.}
To cover complementary forms of logical structure and inference, we first curate a library \(\mathcal{R}\) of formal derivation rule schemata from three logic families: propositional logic, term logic, and first-order logic. Propositional logic captures truth-functional composition and transformations involving logical connectives. Term logic captures categorical and syllogistic relations in the Aristotelian tradition, expressed through relations of inclusion and exclusion between terms. First-order logic provides greater expressive power by introducing variables, predicates, relations, and universal and existential quantifiers. Each schema defines an ordered set of premises and a conclusion that together represent a derivation rule. The library contains 247 schemata: 49 propositional, 77 term, and 121 first-order inference-rule schemata. Representative schemata are shown in Table~\ref{tab:scheme-sample} while a full breakdown is provided in Table~\ref{tab:scheme-family-counts} in the Appendix.

\paragraph{Derivation-Tree Generation.}
Given a target formula $\varphi$ and a library of derivation schemata $\mathcal{R}$, we apply the rules in $\mathcal{R}$ backward to recursively derive a finite sequence of intermediate formulas $\alpha_i$ and a corresponding set of premises $\Gamma$.
Inspired by the LogicTree method~\cite{wang2025logictree}, we construct a derivation tree through backward chaining. Starting from a sampled goal \(\varphi\), we recursively apply compatible derivation rule schemata backwards for at most \(L\) steps. The leaves of the resulting tree constitute the premise set \(\Gamma\), and its root remains the final goal  \(\varphi\). Further dataset details are provided in the    Appendix~\ref{sec:data-details}.

\paragraph{Example.} An example is given in Figure~\ref{fig:pipeline}, where \(Q\lor P\) is selected as the target formula \(\varphi\). Through backward chaining, the generator identifies constructive dilemma as an applicable inference rule and expands the target into three premises: \(G\rightarrow Q\), \(F\rightarrow P\), and \(G\lor F\). The resulting formulas can then be recursively expanded by applying other schemata until the maximum expansion depth is reached or no applicable rule remains.
  
\paragraph{Structure-Preserving Symbol Renaming.} To increase the diversity of the training data and prevent the model from overfitting to simple surface-level patterns, we generate $n$ structure-preserving variants of each derivation tree. Each variant applies injective random renaming to the symbols throughout the tree while preserving the derivations topology, rule sequence, and logical relations.

\paragraph{Next-Step Derivation Construction.} Each derivation is finally expanded into $K$ next-step prediction examples. At derivation step $i$, the input contains the complete set of premises $\Gamma$, the previously derived prefix $(y_1,\ldots,y_{i-1})$, and the final goal $\varphi$. The prediction target is the next valid derivation step $y_i$. The serialized derivations are then adopted for Logic-PPT (Section \ref{sec:logic_ppt}).  

\begin{figure*}[t]
    \centering
    \includegraphics[width=1\linewidth]{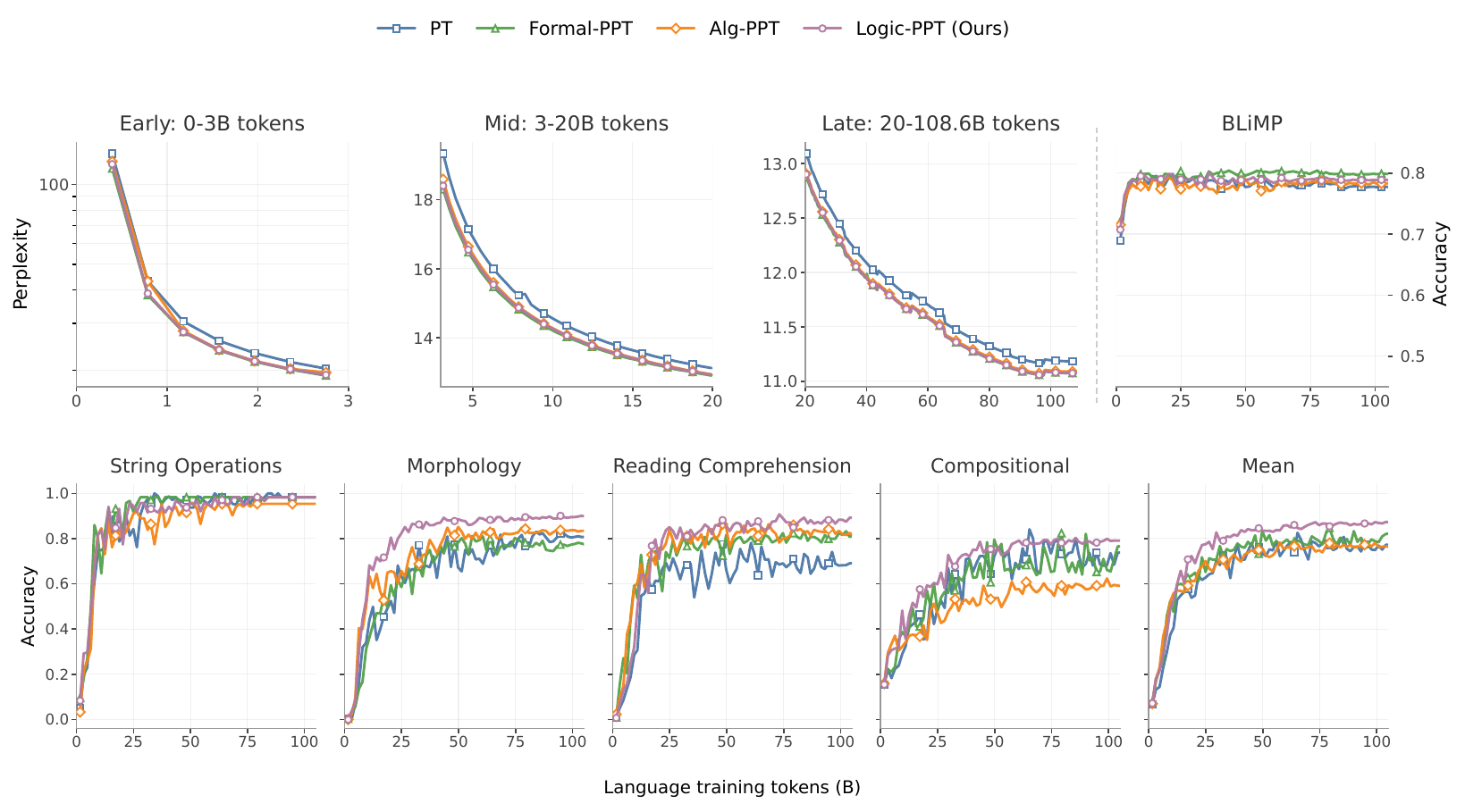}
    \caption{Behavioral trajectories over the 100B-token natural language pretraining run. \textbf{Top:} Language modeling perplexity across three training phases, alongside overall BLiMP accuracy. For BLiMP, Formal-PPT performs slightly better than the other variants, but all four saturate quickly and stay close together. \textbf{Bottom:} Mean accuracy trajectories across elemental tasks \cite{liu2026languagemodelslearnwhen}, shown as an overall mean and broken down by category. Logic-PPT shows substantial improvements on Morphology, Reading Comprehension, and Compositional tasks.}
    \label{fig:4type-acc-trajectories}
\end{figure*}

\section{Experimental Setup}

\paragraph{Baselines.}
We compare \textit{Logic-PPT} against two established symbolic PPT methods using a mixture of procedurally generated algorithmic data (i.e., \textit{Alg-PPT}), including \textsc{Set}, \textsc{Sort}, and \textsc{Union} ~\cite{jiang2026procedural}, and a mixture of formal language data (i.e., \textit{Formal-PPT}) combining \textsc{Dyck} and \textsc{Shuffle-Dyck} \cite{hu-etal-2025-circuits} (see Figure \ref{fig:pipeline}, left). Examples for each class of PPT method are interleaved using balanced round-robin sampling. All PPT methods use an approximately equivalent budget of 2.5B tokens. In addition, we compare our method against a randomly initialized baseline undergoing only language pretraining (i.e., \textit{PT}).

\paragraph{Model Architecture.}

For all configurations, we use a 14-layer Qwen3 architecture~\cite{qwen3technicalreport} as the Transformer backbone. Following the setup of~\citet{yamaguchi-etal-2026-enhancing}, we replace the Qwen3 tokenizer with the 32K Mistral tokenizer~\cite{jiang2023mistral7b} to reduce the compute cost. The resulting model has a 220M-parameter Transformer backbone and 254M parameters in total.

\paragraph{Language Pretraining.}

After PPT and weight transfer, all parameters remain trainable during pretraining on approximately 100B FineWeb-Edu tokens~\cite{lozhkov2024fineweb-edu}. We use LLaMA Factory~\cite{zheng2024llamafactory} with a sequence length of 2,048, a batch size of 96, and checkpoints saved every 2,000 steps to track training progress. Additional implementation and training details, as well as hyperparameters are provided in Table~\ref{tab:100bt-model}, \ref{tab:100bt-hyperparameters}, and \ref{tab:symbolic-pretraining-hyperparameters} in the Appendix.

\paragraph{Evaluating Language and Skill Acquisition.}

We use perplexity on a held-out test set of FineWeb-Edu to evaluate general language modeling capabilities. In addition, to track skill acquisition and emergence during language pretraining, we evaluate the models on BLiMP~\cite{warstadt2020blimp} and on the elemental tasks by ~\citet{liu2026languagemodelslearnwhen}. 

Following ~\citet{liu2026languagemodelslearnwhen}, we retain the elemental tasks for which at least one PPT configuration achieves 80\% accuracy during language pretraining within the 100B token budget. This leads to a total of 17 elemental tasks spanning emergent linguistic capabilities such as string operations, morphology, reading comprehension, and compositional tasks (Table~\ref{tab:emergent-task-categories} in the Appendix).

\section{Results}

\subsection{Language Acquisition}

As shown in Figure~\ref{fig:4type-acc-trajectories}, language modeling perplexity evolves through three training phases. During the early phase (0--3B), the perplexity decreases rapidly under all four configurations. Logic-PPT yields a faster reduction than the PT baseline, producing a clear gap within the first 1B tokens. This advantage persists throughout both the middle (3--20B) and late (20--108.6B) phases. At the final checkpoint, the PT baseline reaches a perplexity of 11.183, compared with 11.088 for Alg-PPT (a 0.85\% reduction) and 11.074 for both Formal-PPT and Logic-PPT (a 0.97\% reduction). These results show that the language modeling advantage of PPT emerges early and remains stable throughout extended natural language training. At the same time, we found that perplexity alone is insufficient to fully characterize the differences between symbolic PPT approaches, with different methods exhibiting comparable trajectories and final perplexity values.

\subsection{Skill Acquisition}

Analysing the results on BLiMP, we observe that all four models improve rapidly early in training, reaching approximately 80\% accuracy within 10-25B tokens, after which performance largely plateaus, with no further gains over the remaining 80B+ tokens. This suggests that, similarly to perplexity, methods relying solely on BLiMP provide little insight into the emergence of more complex capabilities. Despite this, results show that Logic-PPT outperforms the PT baseline and is competitive with alternative symbolic PPT methods.

Performance on the elemental tasks continues to improve throughout the full 100B-token training run, while the separation among training conditions persists, providing deeper insight into skill acquisition dynamics. Across these tasks, the PPT models outperform the PT baseline, indicating that symbolic PPT improves the speed and strength of skill acquisition. 

Logic-PPT, in particular, achieves the 80\% accuracy threshold earlier than the other configurations. By the final checkpoint, the overall mean accuracy across all 17 tasks is 87.4\% for Logic-PPT, versus 80.3\% for Formal-PPT, 77.3\% for PT, and 76.8\% for Alg-PPT. The gains are particularly clear in Morphology, Reading Comprehension, and Compositional tasks, where Logic-PPT outperforms the strongest competing condition by 5.5--6.7\% accuracy points. Further details are in Appendix~\ref{sec:task-performance}.

\section{Mechanistic Analysis}

We analyze model representational dynamics throughout language PT in both activation and weight space. For the activation-space analysis, we use a held-out subset of FineWeb-Edu containing 15K samples and extract the hidden state of the final non-padding token at each layer. We first use centered kernel alignment (CKA)~\cite{kornblith2019similarity} to measure activation similarity both across layers within the same model and between models trained under different PPT configuration. We further compute RankMe~\cite{garrido2023rankme} and the spectral decay~\cite{NEURIPS2022_70596d70} from the centered covariance eigenspectrum of these activations to characterize the geometry and effective complexity. Following \citet{li2025tracing}, we use these metrics to characterize changes in the models' training dynamics. For the weight-space analysis, we compute the stable rank of the model weight matrices to measure how concentrated or distributed their singular-value spectra are. Formal definitions of all metrics are provided in Appendix~\ref{sec:Mechanistic-Analysis-Tools}.

\subsection{Activation Space}

\paragraph{CKA Analysis.}

\begin{figure}[t]
    \centering
    \includegraphics[width=\linewidth]{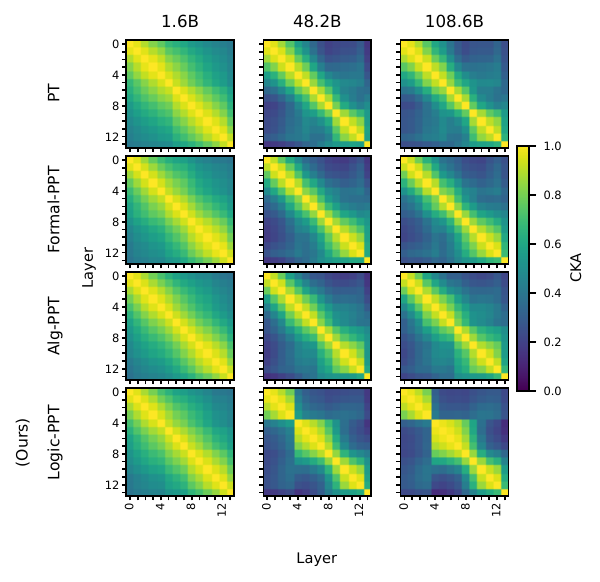}
    \caption{Self layer-to-layer CKA (each model compared against itself) at three checkpoints. Rows are training conditions (PT, Formal-PPT, Alg-PPT, Logic-PPT (Ours)); columns are checkpoints. All four models initially exhibit uniformly high similarity and develop block-diagonal structure as training progresses. However, Logic-PPT exhibits more clear blocks, with lower CKA at the boundary layers.}
    \label{fig:cka}
\end{figure}

\begin{figure}[t]
    \centering
    \includegraphics[width=\linewidth]{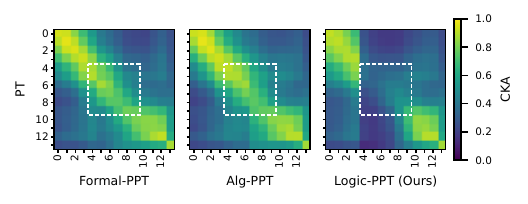}
    \caption{Layer-to-layer CKA between baseline PT and each pre-pretraining (PPT) model, computed at the final checkpoint (108.6B). The $y$-axis indexes PT's layers; the $x$-axis indexes the PPT model's layers. Logic-PPT exhibits lower similarity with PT's  activations, particularly in the middle layers (layers 4--9, marked by the dashed box).}
    \label{fig:pairwise_cross_layer_cka}
\end{figure}

\begin{figure}[t]
    \centering
    \includegraphics[width=0.785\linewidth]{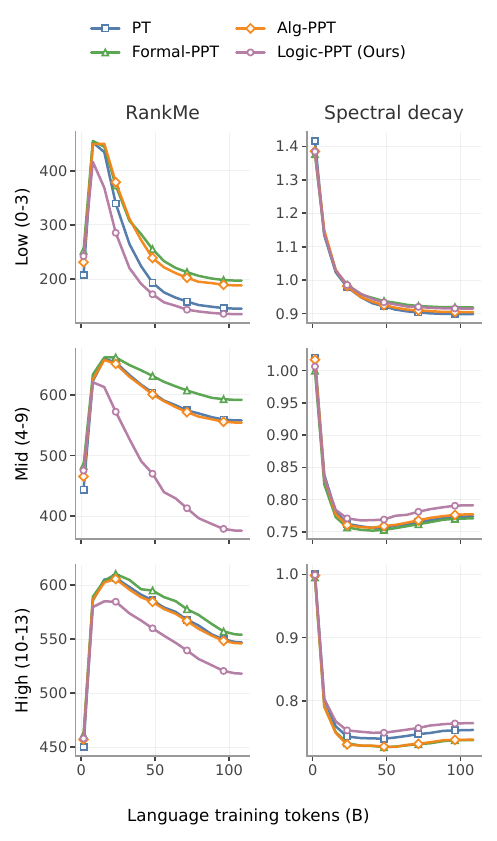}
    \caption{Layer-group trajectories of RankMe and the spectral decay. During the middle and later stages of language PT, Logic-PPT shows much lower RankMe and a higher spectral decay, indicating more compact and lower-dimensional representations.}
    \label{fig:rankme-spectral-decay}
\end{figure}

Figure~\ref{fig:cka} shows that within-model layer-wise CKA becomes increasingly structured during language PT. Early checkpoints exhibit high similarity across many layer pairs, whereas later checkpoints show lower off-diagonal CKA and clearer block structure, indicating increasing layer differentiation and the formation of distinct representational stages. Logic-PPT  produces the clearest block structure, particularly in the middle and upper layers, suggesting a more differentiated and modular organization that persists after extensive language training. Consistent with this finding, pairwise CKA at the final checkpoint (Figure~\ref{fig:pairwise_cross_layer_cka}) shows lower similarity between the Logic-PPT and the PT in the middle layers, indicating that Logic-PPT induces a distinct and persistent activation space organization that fundamentally differs from other symbolic PPT configurations.

\paragraph{RankMe and Spectral Decay.}

 The CKA analysis reveals clear layer-wise representational differences, motivating a closer examination of activation space geometry and effective complexity.
Figure~\ref{fig:rankme-spectral-decay} shows the trajectories of RankMe and the spectral decay in three layer-groups. We observe that early in training RankMe increases while the spectral decay decreases, indicating that variance becomes distributed across a broader set of representational directions. Later, this pattern reverses: RankMe decreases and spectral decay increases, showing that variance is progressively concentrated into fewer dominant directions. Although the transition occurs at different times and with different magnitudes across layers, the overall pattern is consistent with the \emph{entropy-seeking} and \emph{compression-seeking} phases described by~\citet{li2025tracing}. The effects of PPT become most pronounced during this \emph{compression-seeking} phase. Logic-PPT produces the lowest RankMe in this phase and the highest spectral decay in the middle and upper layers, indicating a more compact and anisotropic representation space than the other training configurations. Rather than merely reducing representational dimensionality, Logic-PPT appears to promote a more efficient form of spectral concentration, organizing representations around a smaller set of directions. This interpretation is consistent with our CKA results, which reveal a clearer layer-wise block structure under Logic-PPT.

\subsection{Weight Space}

\begin{figure}[t]
    \centering
    \includegraphics[width=1\linewidth]{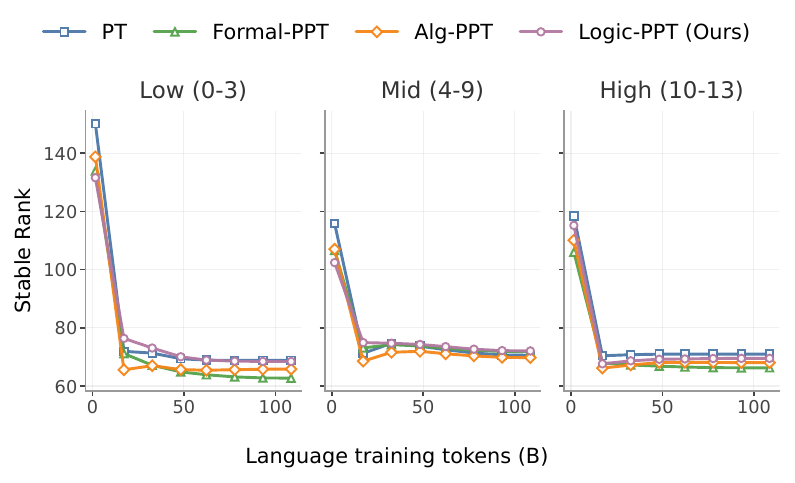}
    \caption{Stable rank trajectories of attention weight matrices across layer groups during language PT. Across all layer groups, PT starts with a higher stable rank than the PPT configurations.}
    \label{fig:stank}
\end{figure}

A complementary analysis of the weight trajectory in Figure~\ref{fig:stank} shows that stable rank drops sharply during the early stage of training and then gradually plateaus. The PPT models generally begin with a lower stable rank than the PT baseline, suggesting that symbolic PPT already biases the attention weights toward a more structured and lower-dimensional spectrum before language exposure.

\begin{figure}[t]
    \centering
    \includegraphics[width=1\linewidth]{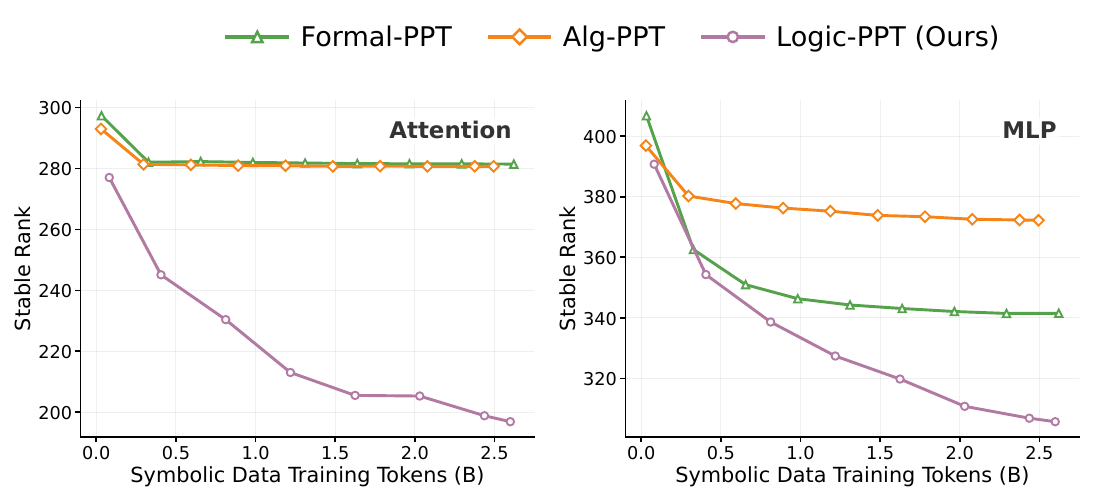}
    \caption{Stable rank of the attention and MLP weight matrices during symbolic PPT. Under Logic-PPT, stable rank continues to decrease, whereas the other PPT conditions plateau.}
    \label{fig:stable-rank-pretraining}
\end{figure}


To further analyse this, Figure~\ref{fig:stable-rank-pretraining} tracks the stable rank of the attention and MLP weight matrices over the PPT stage. Here, under Logic-PPT, stable rank contracts continually throughout training, for both attention and MLP, with no clear plateau. The other symbolic PPT methods, by contrast, show a much weaker effect, in which stable rank drops only in the earliest training steps and quickly flattens out, remaining largely unchanged for the remainder of training. These results suggest that formal derivation data indeed imposes a stronger and more persistent structural bias on the model's weights than the formal-language or algorithmic objectives.

\section{Model Compression}

\begin{figure}[t]
    \centering
    \includegraphics[width=0.875\linewidth]{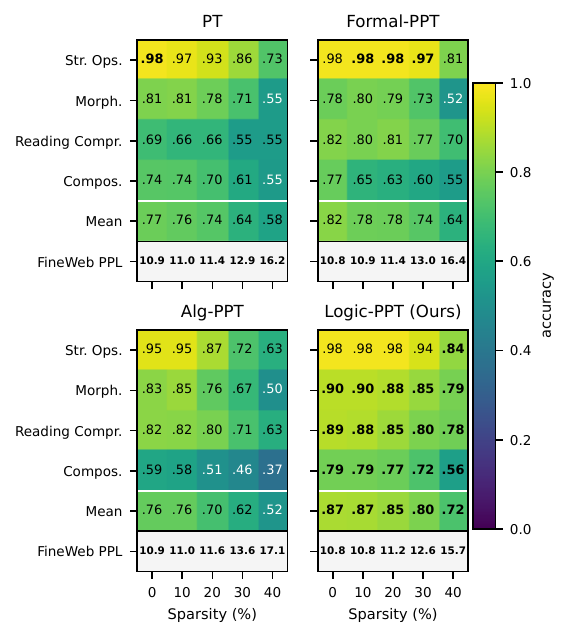}
    \caption{Mean accuracy under Wanda pruning across four elemental task categories, and FineWeb-Edu perplexity under Wanda pruning. Logic-PPT (Ours) performs robustly under compression, retaining the highest accuracy across nearly all categories and sparsity levels.}
    \label{fig:wanda}
\end{figure}

Prior work using CKA has shown that highly similar upper-layer representations permit some layers to be removed with minimal performance degradation in fine-tuned Transformers ~\cite{phang-etal-2021-fine}. Although their experiments concern layer truncation rather than weight pruning, they establish a connection between representational organization and model compressibility. Motivated by this connection, together with our representational and spectral analyses, we hypothesize that Logic-PPT induces a more structured and robust computational organization, enabling the model to better preserve its capabilities under weight pruning. To test this, we apply Wanda pruning~\cite{sun2023wanda} at sparsity levels up to 40\% to models trained under different PPT conditions. Wanda scores weights using both parameter magnitude and input activation strength. We evaluate pruning robustness on the elemental tasks and FineWeb-Edu perplexity. Figure~\ref{fig:wanda} shows that Logic-PPT retains the highest mean accuracy across the four task categories. At 40\% sparsity, Logic-PPT mean accuracy decreases from 0.87 to 0.745, a relative drop of 14.4\%, compared with drops from 24.7\% for the PT baseline, 22.0\% for Formal-PPT, and 31.6\% for Alg-PPT. Logic-PPT also exhibits the smallest increase in FineWeb-Edu perplexity under pruning. Importantly, this robustness advantage does not arise from PPT in general. Formal-PPT and Alg-PPT do not exhibit the same degree of resistance to pruning, with Alg-PPT suffering the largest performance degradation.

\section{Related Work}

\paragraph{Symbolic and Procedural Pre-pretraining.}

Recent work explores whether structured synthetic data can provide useful inductive biases before language pretraining~\cite{mita-etal-2026-language, guo2026syntheticprepretrainingimproveslanguage, lee2026traininglanguagemodelsneural,Shinnick_2026_CVPR,wu2022insights}. \citet{hu-etal-2025-circuits} show that pre-pretraining on formal languages improves language-modeling loss and linguistic generalization, while ~\citet{jiang2026procedural} extend this approach to procedurally generated formal and algorithmic data.  Despite these promising results, prior work
has primarily assessed pre-pretraining through language-modelling loss and downstream accuracy, often at relatively small training scales. 

\paragraph{Logic Data Generation.}

Logic data has been widely used for model evaluation and post-training~\cite{liu2025synlogic,morishita2024enhancing,lacombe2026reasoningcorescalableprocedural,xie2025logicrlunleashingllmreasoning,abdin2024phi4technicalreport,tan-etal-2025-enhancing-logical,kim-etal-2025-reasoning,lee-etal-2026-fol,cao-etal-2026-fundamental}. For example, LogiConBench constructs controllable logical graphs to evaluate model consistency~\cite{chen2026logiconbench}, while LogicTree generates instantiated multi-step reasoning problems with diverse structures and contexts~\cite{wang2025logictree}. In contrast, we use formal derivation before language pretraining to shape representations and inductive biases for broader language learning, rather than to teach specific logic tasks.

\paragraph{Pretraining Dynamics and Mechanistic Analyses.}

Dense checkpoint suites, such as Pythia~\cite{Pythia} and OLMo~\cite{olmo20242olmo2furious}, have enabled researchers to study how language-model capabilities and internal structures evolve throughout pretraining, rather than evaluating only the final checkpoint.
At the behavioural level, language models exhibit broadly consistent skill-acquisition trajectories across random seeds, model families,
and training mixtures~\cite{evanson-etal-2023-language,liu2026languagemodelslearnwhen}. Mechanistic studies reveal systematic changes in representation and weight geometry~\cite{kulkarni2026disentangling,yu2026pretraininginducesreusablespectral,yunis2024approachingdeeplearningspectral}. \citet{li2025tracing} identify warm-up, entropy-seeking, and compression-seeking phases using RankMe and spectral decay, while related work tracks effective rank and spectral structure in Transformer weights~\cite{kulkarni2026disentangling}. 
However, these tools have not yet been used to examine how pre-pretraining reshapes subsequent natural language learning dynamics and internal model organization.

\section{Conclusion}

We investigated how symbolic logic can shape natural language pretraining in LMs, examining its effects from behavioral, mechanistic, and interventional perspectives. Behaviorally, Logic-PPT accelerates the acquisition of linguistic skills, reaching strong task accuracy with substantially fewer natural language PT tokens. Mechanistically, these gains are accompanied by persistent changes in model organization: Logic-PPT develops a more modular, compact, and lower-rank representation space that emerges early and persists throughout training. As an intervention, this altered organization translates into greater robustness under pruning, showing that the structure induced by formal logical derivations yields models that are not only more data-efficient but also more compressible.

\section*{Limitations}

\paragraph{Model Scale and Data.}

Due to computational constraints, we restrict our experiments to a  Qwen architecture with approximately 254M parameters. Following a computationally feasible academic setting like work \cite{yamaguchi-etal-2026-enhancing}, we train each model from scratch on approximately 100B natural language tokens.

\paragraph{Single Seed.}

The main behavioral, representational, and pruning results are based on a single 100B-token training run for each condition. Repeating all pretraining configurations was prohibitive within the constraints of the available academic computing infrastructure. To partially assess seed sensitivity, we train each transferred model with three different initialization seeds for 10B natural language tokens (see Appendix \ref{sec:different-seed}). The resulting self-layer CKA matrices show consistently high correlations across seeds, indicating that the observed layer-wise representational organization is stable at this stage of training.

\paragraph{Evaluation Scope.}

Our evaluation primarily focuses on capability acquisition during language pretraining, using the elemental tasks and BLiMP, together with analyses of representation geometry and pruning robustness. We do not evaluate whether the benefits of Logic-PPT persist after supervised fine-tuning or transfer to a broader range of downstream tasks, such as natural language reasoning, question answering, or instruction following. Future work should examine whether the observed pretraining advantages translate into consistent improvements under different post-training and downstream evaluation settings.

\paragraph{Hyperparameter Selection.}

We approximately match the symbolic PPT conditions by the total number of observed tokens. However, their sequence lengths are task-specific and follow the configurations adopted in prior work, reflecting differences in the structure and serialization of the symbolic tasks.

\paragraph{Language Coverage.}

Our analysis is limited primarily to English-language pretraining and evaluation. We therefore do not test whether symbolic PPT improves multilingual transfer or whether the same skill-acquisition patterns hold across languages with different morphology, syntax, or writing systems.

\section*{Acknowledgments}

We acknowledge the University of Sheffield IT Services for providing access to the Stanage high-performance computing facilities. We also acknowledge the Isambard-AI National AI Research Resource (AIRR), operated by the University of Bristol, for providing computational resources under Award ID 0261-5075-3967-1. This work was further supported by computational resources awarded by the EuroHPC Joint Undertaking under Project ID EHPC-DEV-2026D05-083 through access to the Leonardo supercomputer, hosted by CINECA in Italy.

\bibliography{custom}

\clearpage

\appendix

\section{Mechanistic Analysis Tools}
\label{sec:Mechanistic-Analysis-Tools}
\paragraph{CKA.}
To quantify representational similarity between layers, we use linear centered kernel alignment (CKA)~\cite{kornblith2019similarity}. CKA is invariant to orthogonal transformations and isotropic rescaling, making it well suited for comparing representations across language-model layers~\cite{phang-etal-2021-fine,brown-etal-2023-understanding,gurnee2026verbalizablerepresentationsformglobal}. Let $X_i \in \mathbb{R}^{n \times d_i}$ and $X_j \in \mathbb{R}^{n \times d_j}$ denote the activation matrices extracted from layers $i$ and $j$, where $n$ is the number of examples and $d_i$ and $d_j$ are the corresponding hidden dimensions.

Linear CKA is defined as
  \begin{equation}
  \operatorname{CKA}(X_i,X_j)
  =
  \frac{
      \left\lVert \widetilde{X}_i^{\top}\widetilde{X}_j
      \right\rVert_F^2
  }{
      \left\lVert \widetilde{X}_i^{\top}\widetilde{X}_i
      \right\rVert_F
      \left\lVert \widetilde{X}_j^{\top}\widetilde{X}_j
      \right\rVert_F
  },
  \label{eq:cka}
  \end{equation}
  
where $\widetilde{X}_i = H X_i$ and $\widetilde{X}_j = H X_j$ are column-centered activation matrices, $H = I_n - \frac{1}{n}\mathbf{1}_n\mathbf{1}_n^{\top}$ is the centering matrix, and $\lVert \cdot \rVert_F$ denotes the Frobenius norm.

\paragraph{RankMe and Spectral Decay Rate.}

Following \citet{li2025tracing}, we use RankMe~\cite{garrido2023rankme} and the spectral decay rate~\cite{NEURIPS2022_70596d70} to characterize how different pre-pretraining conditions affect the complexity of the representation space. For each checkpoint and layer, we collect a representation matrix $X \in \mathbb{R}^{n \times d}$ from last-token activations on a 15K-example subset of FineWeb-Edu, where $n$ is the number of examples and $d$ is the hidden dimension. We center the features across examples to obtain $X_c$ and compute the empirical covariance matrix $\hat{\Sigma} = \frac{1}{n} X_c^\top X_c$. Let $\lambda_i$ denote the eigenvalues of $\hat{\Sigma}$. To compute RankMe, we normalize the eigenvalues as $p_i = \frac{\lambda_i}{\sum_j \lambda_j}$ and define RankMe as the exponential entropy of this normalized spectrum:

\begin{equation}
  \operatorname{RankMe}(X)
  =
  \exp\left(-\sum_i p_i \log p_i\right).
  \label{eq:rankme}
  \end{equation}

To measure spectral concentration more directly, we compute the spectral decay rate by fitting a power law to the same covariance eigenspectrum, $\lambda_i \propto i^{-\alpha}$. Equivalently, we fit the linear regression
  \begin{equation}
  \log \lambda_i = c - \alpha \log i .
  \end{equation}
  
The fitted negative slope gives the spectral decay rate $\alpha$. A larger $\alpha$ indicates faster spectral decay, meaning that variance is more strongly concentrated along the leading principal directions. Conversely, a smaller $\alpha$ indicates a flatter spectrum and a more uniform spread of variance across dimensions.

A higher RankMe indicates that representational variance is distributed more evenly across dimensions, corresponding to a higher effective dimensionality. Conversely, a lower RankMe indicates that variance is concentrated in fewer dominant directions, resulting in a more anisotropic and lower-dimensional representation. A larger spectral decay rate indicates a more rapidly decaying spectrum and stronger concentration along the leading principal directions, whereas a smaller spectral decay rate corresponds to a flatter spectrum.

\paragraph{Stable Rank.}
To complement the representation-space analysis, we use stable rank to characterize the effective dimensionality of model weight matrices. For a weight matrix $W$ with singular values $\{\sigma_i\}$, stable rank is defined as
  
  \begin{equation}
  \operatorname{srank}(W)
  =
  \frac{\|W\|_F^2}{\|W\|_2^2}
  =
  \frac{\sum_i \sigma_i^2}{\sigma_1^2}.
  \label{eq:stable-rank}
  \end{equation}
  
Higher stable rank indicates that the weight energy is distributed across more singular directions, while lower stable rank indicates that the matrix is dominated by a few leading directions. We use this metric to test whether different pre-pretraining conditions lead to different levels of spectral concentration in the learned weights.

\section{Pruning}
\label{sec:pruning}
For a weight matrix $\mathbf{W} \in \mathbb{R}^{C_{\mathrm{out}} \times C_{\mathrm{in}}}$ and input activations $\mathbf{X} \in \mathbb{R}^{NL \times C_{\mathrm{in}}}$, Wanda assigns each weight the importance score
\begin{equation}
S_{ij}=|W_{ij}|\left\lVert \mathbf{X}_{j} \right\rVert_2.
\end{equation}
Here, $\mathbf{X}_{j}$ denotes the activation of the $j$-th input channel across all tokens.

\section{Model Architecture}
Table~\ref{tab:100bt-model} summarizes the model architecture used in our experiments.
  \begin{table}[t]
  \centering
  \caption{Model architecture used in the experiments.}
  \label{tab:100bt-model}
  \small
  \begin{tabular}{lc}
  \toprule
  \textbf{Hyperparameter} & \textbf{Qwen3-14L} \\
  \midrule
  \multicolumn{2}{l}{\textit{Model architecture}} \\
  Hidden size                       & 1,024 \\
  Intermediate size                 & 3,072 \\
  Number of hidden layers           & 14 \\
  Number of attention heads         & 16 \\
  Number of key--value heads        & 8 \\
  Attention head dimension          & 128 \\
  Maximum position embeddings       & 40,960 \\
  Maximum window layers             & 14 \\
  RoPE theta                         & 1,000,000 \\
  RMSNorm epsilon                   & $10^{-6}$ \\
  Attention dropout                 & 0.0 \\
  Hidden activation                 & SiLU \\
  Transformer parameters            & 220M \\
  Total parameters                  & 254M \\
 \bottomrule
  \end{tabular}
  \end{table}

\section{Derivation generation details}
\label{sec:data-details}
  Inspired by LogicTree~\cite{wang2025logictree}, we represent each reasoning
  instance as a proof tree constructed through backward chaining. For each
  source example, we select a goal from a predefined pool consisting of manually
  specified seed formulas and the conclusions of the available inference-rule
  schemata. The selected formula becomes the root of the derivation tree.

  Starting from the root, we perform at most \(L\) backward-expansion steps. At
  each step, the generator examines the current leaves and identifies all
  leaf--rule pairs for which the conclusion of the rule can be structurally
  unified with the formula at the leaf. To encourage broad coverage of the rule
  library, it preferentially selects among the applicable rules with the lowest
  cumulative usage counts. One of the corresponding leaf--rule pairs is then
  sampled using a seeded pseudo-random number generator. The substitution
  obtained through unification is applied to the rule, and the selected leaf is
  replaced by the resulting instantiated premises. Schema variables that remain
  unbound after unification are assigned fresh symbols that do not already occur
  in the tree, preventing unintended symbol collisions.

  Expansion terminates after \(L\) rule applications or earlier if no current
  leaf admits a valid backward expansion. The leaves of the completed tree form
  the premise set \(\Gamma=\{p_1,\ldots,p_m\}\), while the root remains the final
  goal \(\varphi\). We then traverse the expanded nodes in post-order to obtain
  a valid leaf-to-root proof sequence
  \begin{equation}
      Y=(y_1,\ldots,y_K),
      \qquad K\leq L,
      \qquad y_K=\varphi.
  \end{equation}
  Because each node is visited only after its children, every step \(y_i\) is
  derivable from the initial premises and the preceding steps
  \(y_1,\ldots,y_{i-1}\). Thus, although the tree is constructed backward from
  the goal, its post-order traversal yields a valid forward derivation from
  \(\Gamma\) to \(\varphi\). We set the maximum number of backward-expansion
  steps to $L=10$ and generate $n=5$ variants for each derivation tree.

\section{Pretraining Details}

After symbolic pre-pretraining, we transfer the Transformer blocks to the language-pretraining stage and reinitialize the token embeddings and language-modeling head to match the natural-language tokenizer. All model parameters remain trainable during language pretraining. We train all model variants on approximately 100B tokens from FineWeb-Edu using the same architecture, data, and optimization configuration; the no-pre-pretraining baseline is trained under identical conditions. Table~\ref{tab:100bt-hyperparameters} summarizes the language-pretraining hyperparameters.

\begin{table}[t]
  \centering
  \caption{Hyperparameters of language-pretraining for the 100B-token experiments.}
  \label{tab:100bt-hyperparameters}
  \small
  \begin{tabular}{@{}p{0.58\columnwidth}p{0.34\columnwidth}@{}}
  \toprule
  \textbf{Hyperparameter} & \textbf{Qwen3-14L} \\
  \midrule
  \multicolumn{2}{l}{\textit{Language pretraining}} \\
  Training corpus                   & FineWeb-Edu \\
  Training tokens                   & $\sim$100B \\
  Tokenizer                         & Mistral \\
  Vocabulary size                   & 32,768 \\
  Sequence length                   & 2,048 \\
  Packing                           & Yes \\
  Batch size per GPU                & 96 \\
  GPU                              & 4$\times$ NVIDIA GH200  \\
  Gradient accumulation steps       & 1 \\
  Training steps                    & 127,160 \\
  Maximum learning rate             & $3\times10^{-4}$ \\
  Learning-rate scheduler           & Cosine \\
  Warmup steps                      & 2,000 \\
  Optimizer                         & AdamW \\
  Adam $\epsilon$                   & $10^{-8}$ \\
  Adam $\beta_1$                    & 0.9 \\
  Adam $\beta_2$                    & 0.999 \\
  Gradient clipping                 & 1.0 \\
  Weight decay                      & 0.01 \\
  Training precision                & BF16 \\
  Initialization seed               & 42 \\
  \bottomrule
  \end{tabular}
  \end{table}

\section{Seed Stability of CKA}
\label{sec:different-seed}
\label{app:cka-seed-stability}

\paragraph{Cross-seed agreement.}
For a given model, let $\mathbf{c}^{(s)} \in \mathbb{R}^{n}$ denote the vector of off-diagonal upper-triangular entries of its self-layer CKA matrix under seed $s$. For each of the three seed pairs $(s_a, s_b)$, we compute

\begin{align*}
\text{MeanAbsDiff}(s_a,s_b)
&=
\frac{1}{n}
\sum_{k=1}^{n}
\left|
c^{(s_a)}_k-c^{(s_b)}_k
\right|,
\\
\text{PearsonR}(s_a,s_b)
&=
\mathrm{corr}\!\left(
\mathbf{c}^{(s_a)},
\mathbf{c}^{(s_b)}
\right).
\end{align*}

The former measures the absolute magnitude of seed-to-seed drift in CKA values; the latter measures whether the relative pattern of which layer pairs are more or less similar is preserved across seeds. 

To examine the effect of random seeds on model representations, we initialize the transferred model with three different seeds and train each run for 10B tokens. As shown in Table~\ref{tab:cka-seed-stability}, the mean absolute difference between CKA matrices is small for all models. At the same time, the mean Pearson correlation between vectorized off-diagonal CKA entries is consistently above $0.99$. This indicates that different seeds produce very similar layer-to-layer representational similarity patterns.

\begin{table}[t]
\centering
\small
\setlength{\tabcolsep}{4pt}
\begin{tabular}{lcc}
\toprule
\textbf{Model} &
\shortstack{\textbf{Mean Abs.}\\\textbf{Diff.}} &
\shortstack{\textbf{Mean Pearson}\\$\boldsymbol{r}$} \\
\midrule
PT & 0.0138 & 0.9955 \\
Formal-PPT    & 0.0201 & 0.9942 \\
Alg-PPT      & 0.0171 & 0.9941 \\
Logic-PPT              & 0.0182 & 0.9926 \\
\bottomrule
\end{tabular}
\caption{Seed stability of self-layer CKA at the 10B-token checkpoint.}
\label{tab:cka-seed-stability}
\end{table}

\section{Licenses}

We adopt the Qwen3 architecture but initialize all model parameters from scratch; no pretrained Qwen3 weights are used. The Mistral-7B-Instruct v0.3 tokenizer is released under the Apache License 2.0. FineWeb-Edu is released under the Open Data Commons Attribution License (ODC-By) v1.0, and its use is also subject to the Common Crawl Terms of Use. The elemental tasks repository is released under the MIT License, while BLiMP is distributed under the Creative Commons Attribution 4.0 License.

\section{Logic schemata}
We organize our formal-derivation schema library into three broad classes: propositional logic, term logic, and first-order logic. Within each class, schemata are further grouped into rule families according to the logical operations or inference patterns they instantiate. As summarized in Table~\ref{tab:scheme-family-counts}, the library contains 49 propositional, 77 term-logic, and 121 first-order schemata, for a total of 247 inference-rule schemata.
\begin{table}[t]
  \centering
  \footnotesize
  \setlength{\tabcolsep}{4pt}
  \caption{
  Counts of derivation schemata in the library by logic type and rule family.
  }
  \label{tab:scheme-family-counts}
  \begin{tabular}{llr}
  \toprule
  \textbf{Logic type}
  & \textbf{Rule family}
  & \textbf{Count} \\
  \midrule

  \multirow{27}{*}{Propositional}
    & De Morgan                    & 4 \\
    & Dilemma                      & 4 \\
    & Distribution                 & 4 \\
    & Resolution                   & 3 \\
    & Absorption                   & 2 \\
    & Association                  & 2 \\
    & Biconditional elimination    & 2 \\
    & Classical laws               & 2 \\
    & Commutation                  & 2 \\
    & Conjunction elimination      & 2 \\
    & Disjunctive syllogism        & 2 \\
    & Double negation              & 2 \\
    & Idempotence                  & 2 \\
    & Material equivalence         & 2 \\
    & Material implication         & 2 \\
    & Transposition                & 2 \\
    & Addition                     & 1 \\
    & Biconditional introduction   & 1 \\
    & Conjunction introduction     & 1 \\
    & Excluded middle              & 1 \\
    & Exportation                  & 1 \\
    & Hypothetical syllogism       & 1 \\
    & Importation                  & 1 \\
    & Modus ponens                 & 1 \\
    & Modus tollens                & 1 \\
    & Proof by cases               & 1 \\
    & \textbf{Total}               & \textbf{49} \\

  \midrule

  \multirow{7}{*}{Term}
    & Categorical syllogisms       & 24 \\
    & Complemented-term variants   & 16 \\
    & Syllogistic reductions       & 15 \\
    & Square of opposition         & 9 \\
    & Sorites chains               & 8 \\
    & Derived syllogisms           & 5 \\
    & \textbf{Total}               & \textbf{77} \\

  \midrule

  \multirow{13}{*}{First-order}
    & Syllogism                    & 57 \\
    & Scope                        & 14 \\
    & Derived patterns             & 12 \\
    & Equality                     & 9 \\
    & Distribution                 & 6 \\
    & Classical quantifier laws    & 4 \\
    & Quantifier negation          & 4 \\
    & Uniqueness                   & 4 \\
    & Vacuous quantification       & 4 \\
    & Quantifier rules             & 3 \\
    & Quantifier commutation       & 2 \\
    & Renaming                     & 2 \\
    & \textbf{Total}               & \textbf{121} \\

  \midrule
  \multicolumn{2}{l}{\textbf{Overall}}
  & \textbf{247} \\
  \bottomrule
  \end{tabular}
  \end{table}

\onecolumn

\section{Logic Data Samples}
\label{sec:data-samples}
\label{tab:data-sample} 

\begingroup
\small
\raggedbottom

\setlength{\abovedisplayskip}{4pt}
\setlength{\belowdisplayskip}{4pt}
\setlength{\abovedisplayshortskip}{3pt}
\setlength{\belowdisplayshortskip}{3pt}

\newcommand{\logicdatastep}[6]{%
  \par\noindent
  \[
  \begin{aligned}
  x^{(#1)}_{#2} ={}&
    \langle\text{\texttt{premises}}\rangle\, #3
    \quad
    \langle\text{\texttt{derived}}\rangle\, #4 \\
  & \langle\text{\texttt{goal}}\rangle\, #5
    \quad
    \langle\text{\texttt{next}}\rangle, \\[-0.25em]
  y^{(#1)}_{#2} ={}& #6
  \end{aligned}
  \]
  \par
}

\subsection{Propositional Logic}

\paragraph{Sample 1}
  \[
  \begin{aligned}
  \Theta &=
  (\neg X\lor A)\land(\neg X\lor B\lor C),\\
  \mathcal{P}^{(1)}
  =
  \{&
  G\lor W;
  W\to X;
  G\to X;
  \neg\Theta\to(\neg\Theta\land\neg Y); \\
  &
  K\to\Theta;
  Y
  \},
  \qquad
  g^{(1)}=A\land(B\lor C).
  \end{aligned}
  \]

\logicdatastep{1}{1}{\mathcal{P}^{(1)}}{\texttt{none}}{g^{(1)}}{W\lor G.}

\logicdatastep{1}{2}{\mathcal{P}^{(1)}}{y^{(1)}_1}{g^{(1)}}{X.}

\logicdatastep{1}{3}{\mathcal{P}^{(1)}}{y^{(1)}_1;y^{(1)}_2}{g^{(1)}}{\neg\Theta\to\neg Y.}

\logicdatastep{1}{4}{\mathcal{P}^{(1)}}{y^{(1)}_1;\cdots;y^{(1)}_3}{g^{(1)}}{Y\to\Theta.}

\logicdatastep{1}{5}{\mathcal{P}^{(1)}}{y^{(1)}_1;\cdots;y^{(1)}_4}{g^{(1)}}{Y\lor K.}

\logicdatastep{1}{6}{\mathcal{P}^{(1)}}{y^{(1)}_1;\cdots;y^{(1)}_5}{g^{(1)}}{\Theta.}

\logicdatastep{1}{7}{\mathcal{P}^{(1)}}{y^{(1)}_1;\cdots;y^{(1)}_6}{g^{(1)}}{\neg X\lor(A\land(B\lor C)).}

\logicdatastep{1}{8}{\mathcal{P}^{(1)}}{y^{(1)}_1;\cdots;y^{(1)}_7}{g^{(1)}}{A\land(B\lor C).}

\logicdatastep{1}{9}{\mathcal{P}^{(1)}}{y^{(1)}_1;\cdots;y^{(1)}_8}{g^{(1)}}{(A\land B)\lor(A\land C).}

\logicdatastep{1}{10}{\mathcal{P}^{(1)}}{y^{(1)}_1;\cdots;y^{(1)}_9}{g^{(1)}}{A\land(B\lor C).}

  \paragraph{Sample 2}
  \par\noindent
  \[
  \begin{aligned}
  \mathcal{P}^{(2)}
  =
  \{&
  O;
  A\to L;
  L\to M;
  (M\to B\lor C)\to D;\\
  &
  D\to(M\to B\lor C);
  D
  \},
  \qquad
  g^{(2)}=A\to B\lor C.
  \end{aligned}
  \]

  \logicdatastep{2}{1}{\mathcal{P}^{(2)}}{\texttt{none}}{g^{(2)}}{\neg D\to\neg(M\to B\lor C).}
\logicdatastep{2}{2}{\mathcal{P}^{(2)}}{y^{(2)}_1}{g^{(2)}}{(M\to B\lor C)\to D.}
\logicdatastep{2}{3}{\mathcal{P}^{(2)}}{y^{(2)}_1;y^{(2)}_2}{g^{(2)}}{D\land M\to B\lor C.}
\logicdatastep{2}{4}{\mathcal{P}^{(2)}}{y^{(2)}_1;\cdots;y^{(2)}_3}{g^{(2)}}{D\to(M\to B\lor C).}
\logicdatastep{2}{5}{\mathcal{P}^{(2)}}{y^{(2)}_1;\cdots;y^{(2)}_4}{g^{(2)}}{(M\to B\lor C)\leftrightarrow D.}
\logicdatastep{2}{6}{\mathcal{P}^{(2)}}{y^{(2)}_1;\cdots;y^{(2)}_5}{g^{(2)}}{M\to B\lor C.}
\logicdatastep{2}{7}{\mathcal{P}^{(2)}}{y^{(2)}_1;\cdots;y^{(2)}_6}{g^{(2)}}{L\to B\lor C.}
\logicdatastep{2}{8}{\mathcal{P}^{(2)}}{y^{(2)}_1;\cdots;y^{(2)}_7}{g^{(2)}}{A\to B\lor C.}
\logicdatastep{2}{9}{\mathcal{P}^{(2)}}{y^{(2)}_1;\cdots;y^{(2)}_8}{g^{(2)}}{O\land(A\to B\lor C).}
\logicdatastep{2}{10}{\mathcal{P}^{(2)}}{y^{(2)}_1;\cdots;y^{(2)}_9}{g^{(2)}}{A\to B\lor C.}

  \paragraph{Sample 3}
  \[
  \begin{aligned}
  \mathcal{P}^{(3)}
  =
  \{&
  L\land(\neg B\lor C);
  A
  \},
  \qquad
  g^{(3)}=A\land(B\to C).
  \end{aligned}
  \]

  \logicdatastep{3}{1}{\mathcal{P}^{(3)}}{\texttt{none}}{g^{(3)}}{(L\land\neg B)\lor(L\land C).}
\logicdatastep{3}{2}{\mathcal{P}^{(3)}}{y^{(3)}_1}{g^{(3)}}{L\land(\neg B\lor C).}
\logicdatastep{3}{3}{\mathcal{P}^{(3)}}{y^{(3)}_1;y^{(3)}_2}{g^{(3)}}{(L\land\neg B)\lor(L\land C).}
\logicdatastep{3}{4}{\mathcal{P}^{(3)}}{y^{(3)}_1;\cdots;y^{(3)}_3}{g^{(3)}}{((L\land\neg B)\lor L)\land((L\land\neg B)\lor C).}
\logicdatastep{3}{5}{\mathcal{P}^{(3)}}{y^{(3)}_1;\cdots;y^{(3)}_4}{g^{(3)}}{(L\land\neg B)\lor(L\land C).}
\logicdatastep{3}{6}{\mathcal{P}^{(3)}}{y^{(3)}_1;\cdots;y^{(3)}_5}{g^{(3)}}{L\land(\neg B\lor C).}
\logicdatastep{3}{7}{\mathcal{P}^{(3)}}{y^{(3)}_1;\cdots;y^{(3)}_6}{g^{(3)}}{\neg B\lor C.}
\logicdatastep{3}{8}{\mathcal{P}^{(3)}}{y^{(3)}_1;\cdots;y^{(3)}_7}{g^{(3)}}{B\to C.}
\logicdatastep{3}{9}{\mathcal{P}^{(3)}}{y^{(3)}_1;\cdots;y^{(3)}_8}{g^{(3)}}{(B\to C)\land A.}
\logicdatastep{3}{10}{\mathcal{P}^{(3)}}{y^{(3)}_1;\cdots;y^{(3)}_9}{g^{(3)}}{A\land(B\to C).}

\subsection{Term Logic}

\paragraph{Sample 1}
  \[
  \begin{aligned}
  \mathcal{P}^{(1)}
  =
  \{&
  A\models T;
  T\models V;
  V\models J;
  J\models F;
  F\models \neg O;
  \neg O\models X; \\
  &
  X\models \neg N;
  \neg N\models H;
  H\models Z;
  Z\models \neg Q;
  \neg Q\models M;
  M\models B
  \},
  \\
  g^{(1)}&=A\models B.
  \end{aligned}
  \]

  \logicdatastep{1}{1}{\mathcal{P}^{(1)}}{\texttt{none}}{g^{(1)}}{J\models \neg O.}
\logicdatastep{1}{2}{\mathcal{P}^{(1)}}{y^{(1)}_1}{g^{(1)}}{\neg O\models \neg N.}
\logicdatastep{1}{3}{\mathcal{P}^{(1)}}{y^{(1)}_1;y^{(1)}_2}{g^{(1)}}{\neg O\models H.}
\logicdatastep{1}{4}{\mathcal{P}^{(1)}}{y^{(1)}_1;\cdots;y^{(1)}_3}{g^{(1)}}{Q\models \neg H.}
\logicdatastep{1}{5}{\mathcal{P}^{(1)}}{y^{(1)}_1;\cdots;y^{(1)}_4}{g^{(1)}}{\neg O\models \neg Q.}
\logicdatastep{1}{6}{\mathcal{P}^{(1)}}{y^{(1)}_1;\cdots;y^{(1)}_5}{g^{(1)}}{J\models \neg Q.}
\logicdatastep{1}{7}{\mathcal{P}^{(1)}}{y^{(1)}_1;\cdots;y^{(1)}_6}{g^{(1)}}{J\models M.}
\logicdatastep{1}{8}{\mathcal{P}^{(1)}}{y^{(1)}_1;\cdots;y^{(1)}_7}{g^{(1)}}{V\models M.}
\logicdatastep{1}{9}{\mathcal{P}^{(1)}}{y^{(1)}_1;\cdots;y^{(1)}_8}{g^{(1)}}{T\models M.}
\logicdatastep{1}{10}{\mathcal{P}^{(1)}}{y^{(1)}_1;\cdots;y^{(1)}_9}{g^{(1)}}{A\models B.}

  \paragraph{Sample 2}
  \[
  \begin{aligned}
  \mathcal{P}^{(2)}
  =
  \{&
  A\models G;
  G\models \neg M;
  \neg M\models R;
  R\models U; \\
  &
  \neg C\models E;
  E\models \neg Q;
  \neg Q\models X;
  X\models \neg W;
  \neg W\models \neg U
  \},
  \\
  g^{(2)}&=A\not\models \neg C.
  \end{aligned}
  \]

  \logicdatastep{2}{1}{\mathcal{P}^{(2)}}{\texttt{none}}{g^{(2)}}{G\models R.}
\logicdatastep{2}{2}{\mathcal{P}^{(2)}}{y^{(2)}_1}{g^{(2)}}{\neg C\models \neg Q.}
\logicdatastep{2}{3}{\mathcal{P}^{(2)}}{y^{(2)}_1;y^{(2)}_2}{g^{(2)}}{\neg C\models X.}
\logicdatastep{2}{4}{\mathcal{P}^{(2)}}{y^{(2)}_1;\cdots;y^{(2)}_3}{g^{(2)}}{X\models \neg U.}
\logicdatastep{2}{5}{\mathcal{P}^{(2)}}{y^{(2)}_1;\cdots;y^{(2)}_4}{g^{(2)}}{\neg C\models \neg U.}
\logicdatastep{2}{6}{\mathcal{P}^{(2)}}{y^{(2)}_1;\cdots;y^{(2)}_5}{g^{(2)}}{R\models \neg\neg C.}
\logicdatastep{2}{7}{\mathcal{P}^{(2)}}{y^{(2)}_1;\cdots;y^{(2)}_6}{g^{(2)}}{G\models \neg\neg C.}
\logicdatastep{2}{8}{\mathcal{P}^{(2)}}{y^{(2)}_1;\cdots;y^{(2)}_7}{g^{(2)}}{A\models \neg\neg C.}
\logicdatastep{2}{9}{\mathcal{P}^{(2)}}{y^{(2)}_1;\cdots;y^{(2)}_8}{g^{(2)}}{\neg(A\not\models \neg\neg C).}
\logicdatastep{2}{10}{\mathcal{P}^{(2)}}{y^{(2)}_1;\cdots;y^{(2)}_9}{g^{(2)}}{A\not\models \neg C.}

  \paragraph{Sample 3}
  \[
  \begin{aligned}
  \mathcal{P}^{(3)}
  =
  \{&
  Y\models F;
  F\models L;
  A\models \neg L;
  \neg Y\models D; \\
  &
  D\models K;
  K\models \neg T;
  \neg T\models W;
  W\models B
  \},
  \\
  g^{(3)}&=\neg(A\models \neg B).
  \end{aligned}
  \]

  \logicdatastep{3}{1}{\mathcal{P}^{(3)}}{\texttt{none}}{g^{(3)}}{Y\models L.}
\logicdatastep{3}{2}{\mathcal{P}^{(3)}}{y^{(3)}_1}{g^{(3)}}{A\models \neg Y.}
\logicdatastep{3}{3}{\mathcal{P}^{(3)}}{y^{(3)}_1;y^{(3)}_2}{g^{(3)}}{\neg Y\models K.}
\logicdatastep{3}{4}{\mathcal{P}^{(3)}}{y^{(3)}_1;\cdots;y^{(3)}_3}{g^{(3)}}{\neg Y\models \neg T.}
\logicdatastep{3}{5}{\mathcal{P}^{(3)}}{y^{(3)}_1;\cdots;y^{(3)}_4}{g^{(3)}}{\neg Y\models W.}
\logicdatastep{3}{6}{\mathcal{P}^{(3)}}{y^{(3)}_1;\cdots;y^{(3)}_5}{g^{(3)}}{\neg Y\models B.}
\logicdatastep{3}{7}{\mathcal{P}^{(3)}}{y^{(3)}_1;\cdots;y^{(3)}_6}{g^{(3)}}{A\models B.}
\logicdatastep{3}{8}{\mathcal{P}^{(3)}}{y^{(3)}_1;\cdots;y^{(3)}_7}{g^{(3)}}{\neg(A\not\models B).}
\logicdatastep{3}{9}{\mathcal{P}^{(3)}}{y^{(3)}_1;\cdots;y^{(3)}_8}{g^{(3)}}{A\not\models \neg B.}
\logicdatastep{3}{10}{\mathcal{P}^{(3)}}{y^{(3)}_1;\cdots;y^{(3)}_9}{g^{(3)}}{\neg(A\models \neg B).}

\subsection{First-Order Logic}

\paragraph{Sample 1}
  \[
  \begin{aligned}
  \mathcal{P}^{(1)}
  =
  \{&
  \forall x\, (N(x)\to K(x));
  \forall x\, (K(x)\to O(x));
  \forall x\, (O(x)\to \Psi(x)); \\
  &
  \forall x\, I(x)\lor \forall x\, N(x);
  \forall x\, \neg I(x)
  \},
  \qquad
  g^{(1)}=\forall x\, \Psi(x).
  \end{aligned}
  \]

  \logicdatastep{1}{1}{\mathcal{P}^{(1)}}{\texttt{none}}{g^{(1)}}{\forall x\, (K(x)\to O(x)).}
\logicdatastep{1}{2}{\mathcal{P}^{(1)}}{y^{(1)}_1}{g^{(1)}}{\forall x\, (N(x)\to \Psi(x)).}
\logicdatastep{1}{3}{\mathcal{P}^{(1)}}{y^{(1)}_1;y^{(1)}_2}{g^{(1)}}{\forall x\, I(x)\lor \forall x\, N(x).}
\logicdatastep{1}{4}{\mathcal{P}^{(1)}}{y^{(1)}_1;\cdots;y^{(1)}_3}{g^{(1)}}{\forall x\, (I(x)\lor N(x)).}
\logicdatastep{1}{5}{\mathcal{P}^{(1)}}{y^{(1)}_1;\cdots;y^{(1)}_4}{g^{(1)}}{\neg\exists x\, I(x).}
\logicdatastep{1}{6}{\mathcal{P}^{(1)}}{y^{(1)}_1;\cdots;y^{(1)}_5}{g^{(1)}}{\forall x\, \neg I(x).}
\logicdatastep{1}{7}{\mathcal{P}^{(1)}}{y^{(1)}_1;\cdots;y^{(1)}_6}{g^{(1)}}{\forall x\, N(x).}
\logicdatastep{1}{8}{\mathcal{P}^{(1)}}{y^{(1)}_1;\cdots;y^{(1)}_7}{g^{(1)}}{\neg\exists x\, \neg N(x).}
\logicdatastep{1}{9}{\mathcal{P}^{(1)}}{y^{(1)}_1;\cdots;y^{(1)}_8}{g^{(1)}}{\forall x\, N(x).}
\logicdatastep{1}{10}{\mathcal{P}^{(1)}}{y^{(1)}_1;\cdots;y^{(1)}_9}{g^{(1)}}{\forall x\, \Psi(x).}
  
  \paragraph{Sample 2}
  \[
  \begin{aligned}
  \mathcal{P}^{(2)}
  =
  \{&
  \forall x\, X(x)\lor \forall x\, \Phi(x);
  \neg\exists x\, X(x)
  \},
  \qquad
  g^{(2)}=\neg\exists x\, \neg\Phi(x).
  \end{aligned}
  \]

  \logicdatastep{2}{1}{\mathcal{P}^{(2)}}{\texttt{none}}{g^{(2)}}{\forall x\, (X(x)\lor \Phi(x)).}
\logicdatastep{2}{2}{\mathcal{P}^{(2)}}{y^{(2)}_1}{g^{(2)}}{\forall x\, X(x)\lor \Phi(x).}
\logicdatastep{2}{3}{\mathcal{P}^{(2)}}{y^{(2)}_1;y^{(2)}_2}{g^{(2)}}{\forall x\, (X(x)\lor \Phi(x)).}
\logicdatastep{2}{4}{\mathcal{P}^{(2)}}{y^{(2)}_1;\cdots;y^{(2)}_3}{g^{(2)}}{\forall x\, X(x)\lor \Phi(x).}
\logicdatastep{2}{5}{\mathcal{P}^{(2)}}{y^{(2)}_1;\cdots;y^{(2)}_4}{g^{(2)}}{\forall x\, (X(x)\lor \Phi(x)).}
\logicdatastep{2}{6}{\mathcal{P}^{(2)}}{y^{(2)}_1;\cdots;y^{(2)}_5}{g^{(2)}}{\forall x\, \neg X(x).}
\logicdatastep{2}{7}{\mathcal{P}^{(2)}}{y^{(2)}_1;\cdots;y^{(2)}_6}{g^{(2)}}{\neg\exists x\, X(x).}
\logicdatastep{2}{8}{\mathcal{P}^{(2)}}{y^{(2)}_1;\cdots;y^{(2)}_7}{g^{(2)}}{\forall x\, \neg X(x).}
\logicdatastep{2}{9}{\mathcal{P}^{(2)}}{y^{(2)}_1;\cdots;y^{(2)}_8}{g^{(2)}}{\forall x\, \Phi(x).}
\logicdatastep{2}{10}{\mathcal{P}^{(2)}}{y^{(2)}_1;\cdots;y^{(2)}_9}{g^{(2)}}{\neg\exists x\, \neg\Phi(x).}
  
  \paragraph{Sample 3}
  \[
  \begin{aligned}
  \mathcal{P}^{(3)}
  =
  \{&
  \forall z\, (I(z)\to C(z));
  \forall z\, (C(z)\to E(z));
  \forall z\, (L(z)\to I(z)); \\
  &
  \exists z\, L(z);
  \forall z\, \forall r\, ((E(z)\land E(r))\to z=r); \\
  &
  c=w;
  s=w;
  \forall l\, (A(l)\to E(l));
  A(c);
  E(t)
  \},
  \qquad
  g^{(3)}=s=t.
  \end{aligned}
  \]

  \logicdatastep{3}{1}{\mathcal{P}^{(3)}}{\texttt{none}}{g^{(3)}}{\exists z\, I(z).}
\logicdatastep{3}{2}{\mathcal{P}^{(3)}}{y^{(3)}_1}{g^{(3)}}{\exists z\, E(z).}
\logicdatastep{3}{3}{\mathcal{P}^{(3)}}{y^{(3)}_1;y^{(3)}_2}{g^{(3)}}{\exists!z\, E(z).}
\logicdatastep{3}{4}{\mathcal{P}^{(3)}}{y^{(3)}_1;\cdots;y^{(3)}_3}{g^{(3)}}{\exists z\, (E(z)\land \forall p\, (E(p)\to p=z)).}
\logicdatastep{3}{5}{\mathcal{P}^{(3)}}{y^{(3)}_1;\cdots;y^{(3)}_4}{g^{(3)}}{\exists!z\, E(z).}
\logicdatastep{3}{6}{\mathcal{P}^{(3)}}{y^{(3)}_1;\cdots;y^{(3)}_5}{g^{(3)}}{w=s.}
\logicdatastep{3}{7}{\mathcal{P}^{(3)}}{y^{(3)}_1;\cdots;y^{(3)}_6}{g^{(3)}}{c=s.}
\logicdatastep{3}{8}{\mathcal{P}^{(3)}}{y^{(3)}_1;\cdots;y^{(3)}_7}{g^{(3)}}{E(c).}
\logicdatastep{3}{9}{\mathcal{P}^{(3)}}{y^{(3)}_1;\cdots;y^{(3)}_8}{g^{(3)}}{E(s).}
\logicdatastep{3}{10}{\mathcal{P}^{(3)}}{y^{(3)}_1;\cdots;y^{(3)}_9}{g^{(3)}}{s=t.}

  \endgroup

\section{Pre-pretraining Details}

\begin{table*}[!ht]
  \centering
  \caption{Hyperparameters for the three symbolic PPT conditions. All models are initialized from scratch with the same seed.}
  \label{tab:symbolic-pretraining-hyperparameters}
  \small
  \setlength{\tabcolsep}{5pt}
  \begin{tabular}{lccc}
  \toprule
  \textbf{Hyperparameter}
  & \textbf{Logic}
  & \textbf{Algorithmic}
  & \textbf{Formal Language} \\
    \midrule
  Training tasks
  & Prop./FOL/term logic
  & Set, sort, union
  & Dyck, Shuffle-Dyck \\
  Task sampling
  & Mixed dataset
  & Balanced round-robin
  & Balanced round-robin \\
  Tokenizer
  & Character-level
  & Task-specific discrete
  & Task-specific discrete \\
  Vocabulary size
  & 74 & 231 & 359 \\
  Training examples
  & 6.0M total
  & 6.0M per task
  & 20.5M per task \\
  Maximum sequence length
  & 2,048 & 128 & 128 \\
  Batch size
  & 256 & 256 & 256 \\
  Gradient accumulation
  & 1 & 1 & 1 \\
  Selected checkpoint step
  & 32,000 & 84,000 & 80,000 \\
  Input tokens observed
  & 2.600B & 2.753B & 2.621B \\
  Maximum learning rate
  & $1\times10^{-4}$
  & $5\times10^{-5}$
  & $5\times10^{-5}$ \\
  Learning-rate schedule
  & Constant
  & Constant
  & Constant \\
  Warmup steps
  & 100 & 100 & 100 \\
  Optimizer
  & AdamW & AdamW & AdamW \\
  Adam $\epsilon$
  & $10^{-8}$ & $10^{-8}$ & $10^{-8}$ \\
  Adam $\beta_1$
  & 0.9 & 0.9 & 0.9 \\
  Adam $\beta_2$
  & 0.999 & 0.999 & 0.999 \\
  Weight decay
  & 0.01 & 0.1 & 0.1 \\
  Gradient clipping
  & 1.0 & 1.0 & 1.0 \\
  Training precision
  & BF16 & BF16 & BF16 \\
  Loss positions
  & Target only
  & Target only
  & All non-final tokens \\
  Initialization seed
  & 42 & 42 & 42 \\
  GPU
    & \multicolumn{3}{c}{1$\times$ NVIDIA GH200 (96 GB)} \\

  \bottomrule
  \end{tabular}
  \end{table*}

\twocolumn

\section{Task Performance}
\label{sec:task-performance}
\begin{figure*}
    \centering
    \includegraphics[width=1\linewidth]{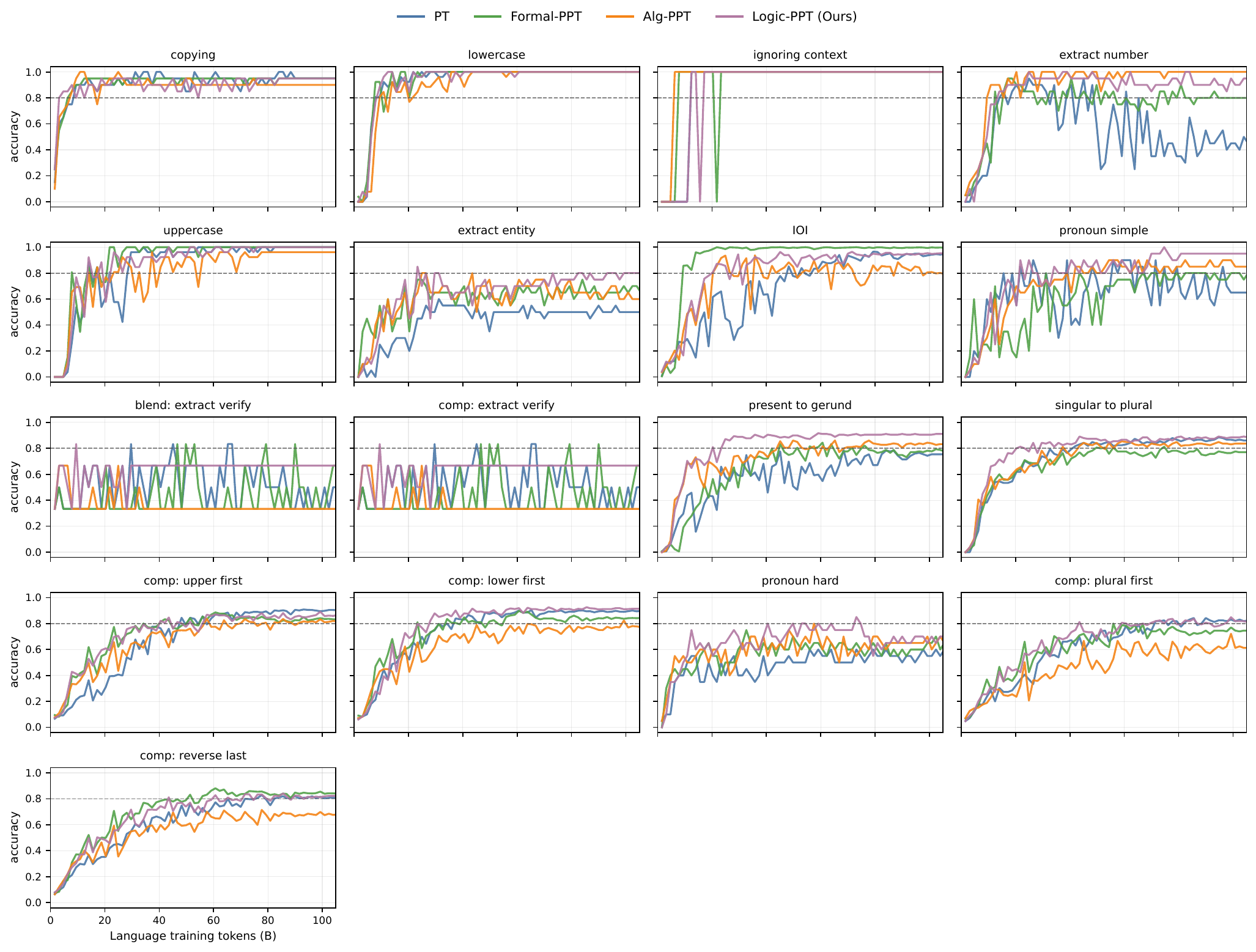}
    \caption{Accuracy trajectories for the 17 elemental tasks across language training tokens.}
    \label{fig:elemental-task-accuracy-curves}
\end{figure*}

\begin{figure*}
    \centering
    \includegraphics[width=1\linewidth]{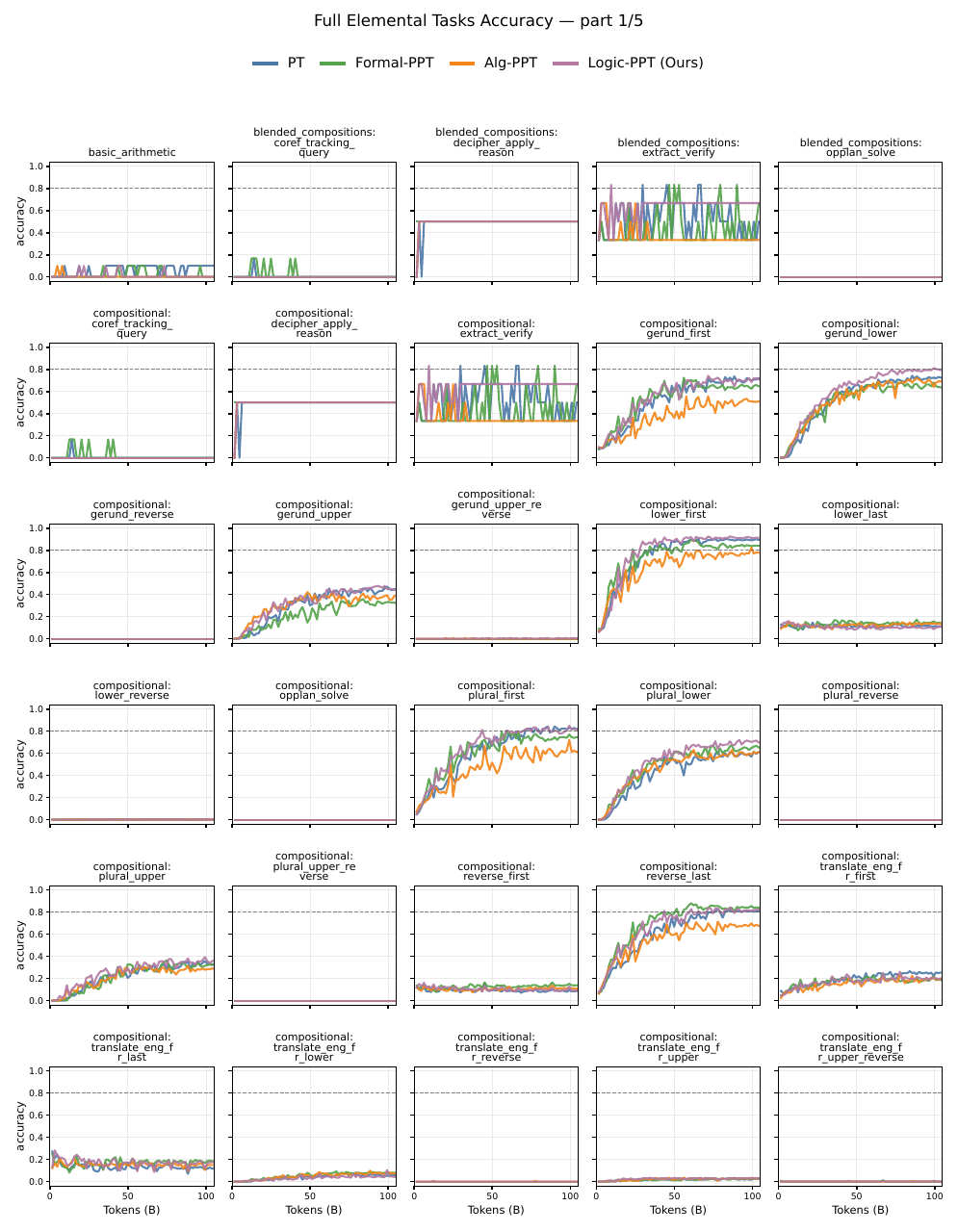}
    \caption{Accuracy trajectories for the first subset of tasks in the full 130-task ElementalTask suite.}
    \label{fig:accuracy_curves_1}
\end{figure*}

\begin{figure*}
    \centering
    \includegraphics[width=1\linewidth]{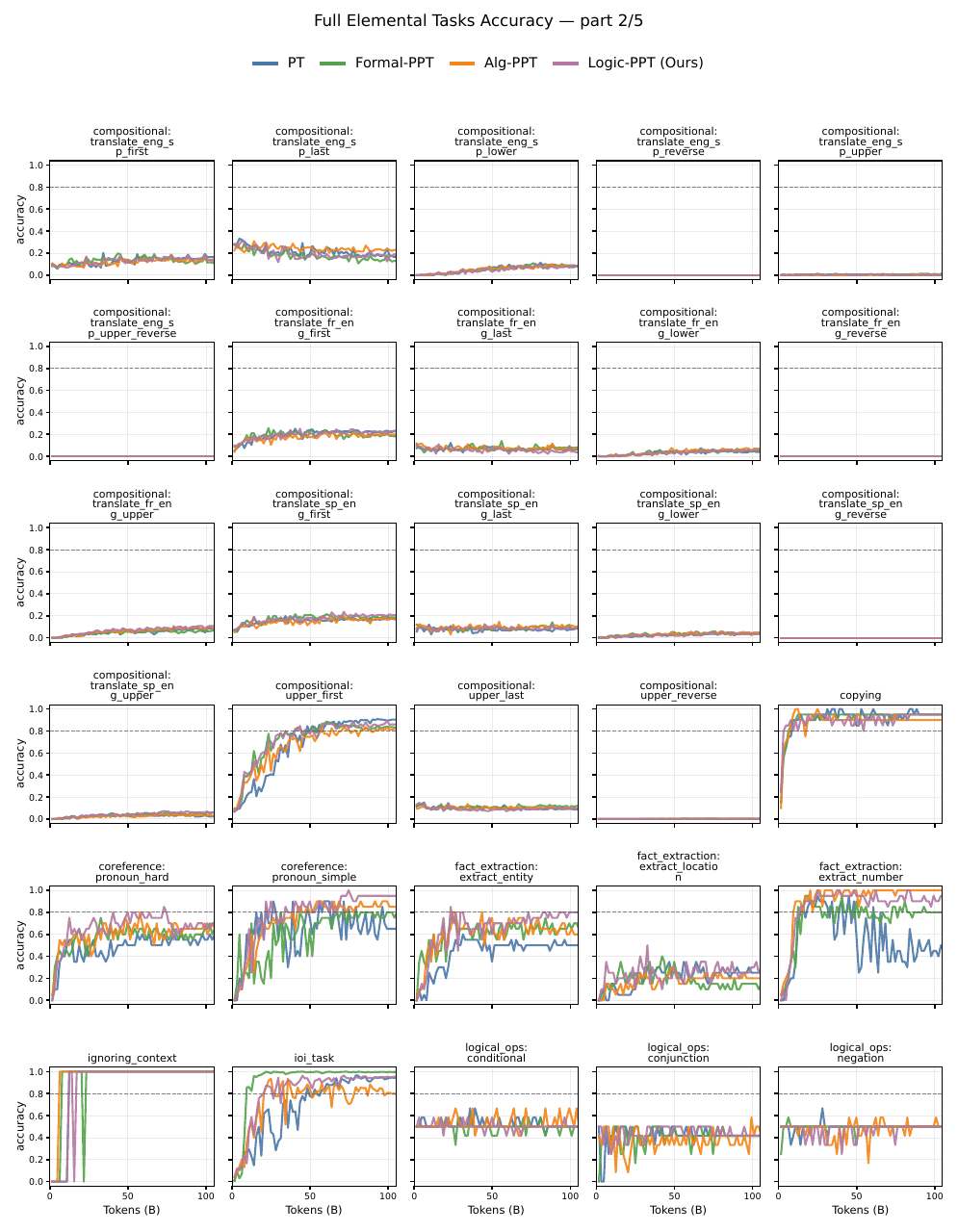}
    \caption{Accuracy trajectories for the second subset of tasks in the full 130-task ElementalTask suite.}
    \label{fig:accuracy_curves_2}
\end{figure*}

\begin{figure*}
    \centering
    \includegraphics[width=1\linewidth]{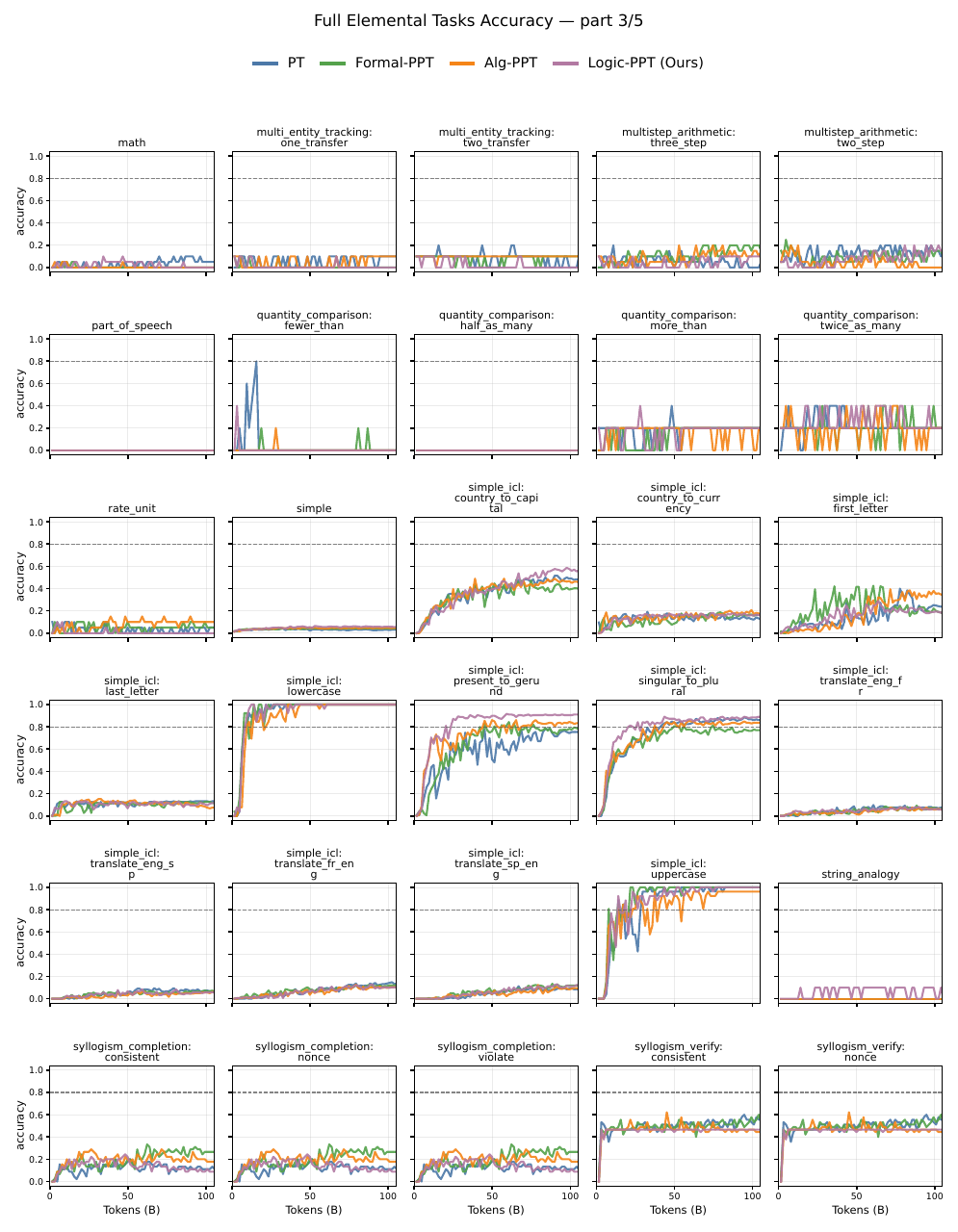}
    \caption{Accuracy trajectories for the third subset of tasks in the full 130-task ElementalTask suite.}
    \label{fig:accuracy_curves_3}
\end{figure*}

\begin{figure*}
    \centering
    \includegraphics[width=1\linewidth]{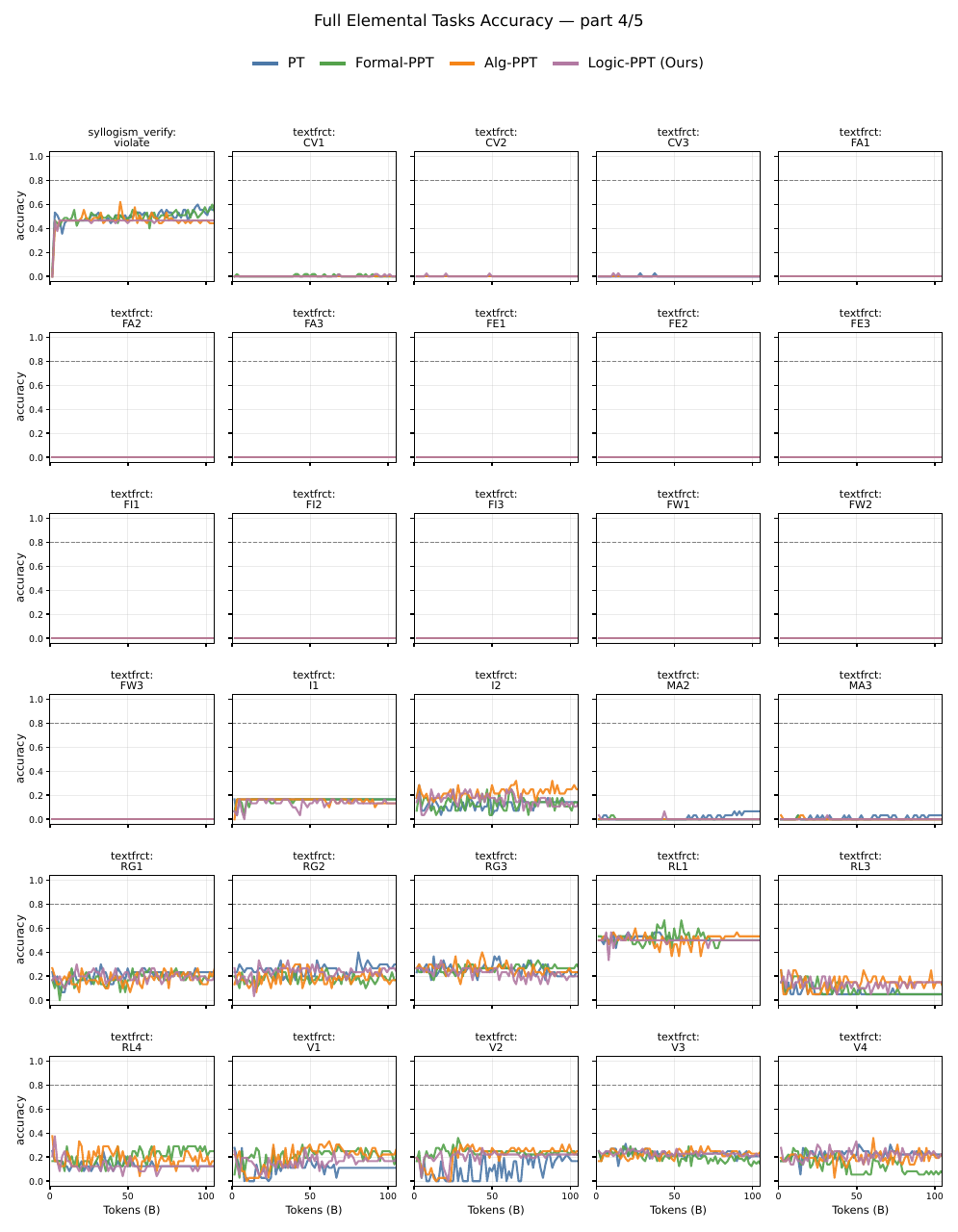}
    \caption{Accuracy trajectories for the fourth subset of tasks in the full 130-task ElementalTask suite.}
    \label{fig:accuracy_curves_4}
\end{figure*}

\begin{figure*}
    \centering
    \includegraphics[width=1\linewidth]{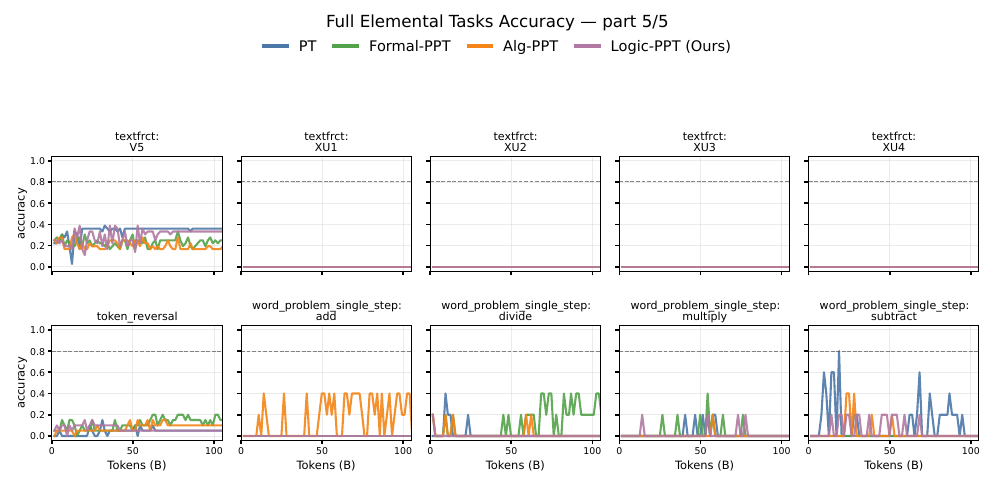}
    \caption{Accuracy trajectories for the fifth subset of tasks in the full 130-task ElementalTask suite.}
    \label{fig:accuracy_curves_5}
\end{figure*}

\begin{figure*}
    \centering
    \includegraphics[width=1\linewidth]{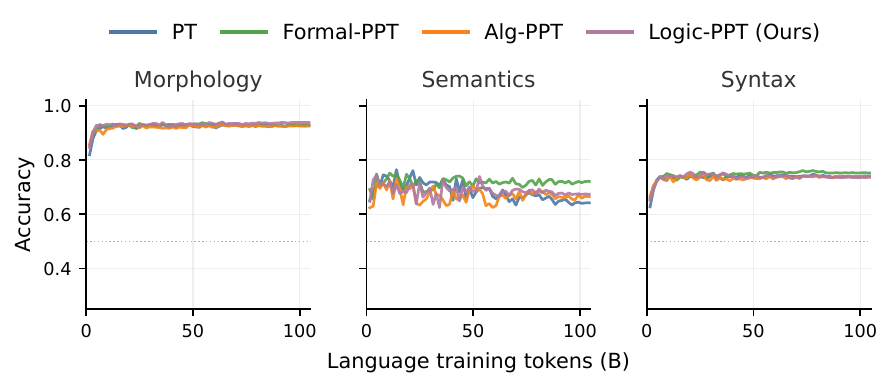}
    \caption{BLiMP accuracy trajectories by linguistic field.}
    \label{fig:blimp}
\end{figure*}

We report full accuracy trajectories on the ElementalTask suite and BLiMP accuracy broken down by linguistic field, to complement the aggregate results discussed above with per-task and per-field detail. Figure~\ref{fig:elemental-task-accuracy-curves} shows accuracy versus cumulative language training tokens for the 17 elemental tasks used in our main emergence analysis, grouped into the four skill categories introduced earlier---String Operations, Morphology, Reading Comprehension, and Compositional---for all four model variants (PT, Formal-PPT, Alg-PPT, Logic-PPT). To check that this curated subset is representative rather than cherry-picked, Figures~\ref{fig:accuracy_curves_1}--\ref{fig:accuracy_curves_5} report the same trajectories for the full 130-task elemental task suite, split across five panels for legibility; the 17 tasks in Figure~\ref{fig:elemental-task-accuracy-curves} are the subset without a clean emergence point, used for the emergence-order analysis. Figure~\ref{fig:blimp} further decomposes the overall BLiMP trajectory by field (Morphology, Syntax, Semantics), showing that the aggregate curves reported earlier mask substantial field-level variation. Table~\ref{tab:spearman} reports the pairwise Spearman rank correlations between task-emergence orders for every pair of models, providing the data underlying the implicit-curriculum discussion.

\begin{table}[!t]
\centering
\small
\begin{tabular}{lccc}
\toprule
\textbf{Pair} & \textbf{$\rho$} & \textbf{p-value} & \textbf{n} \\
\midrule
PT vs Formal-PPT & 0.800 & 0.00058 & 14 \\
PT vs Alg-PPT & 0.768 & 0.00954 & 10 \\
PT vs Logic-PPT & \textbf{0.801} & 0.00057 & 14 \\
Formal-PPT vs Alg-PPT & 0.768 & 0.00579 & 11 \\
Formal-PPT vs Logic-PPT & 0.557 & 0.03084 & 15 \\
Algo-PPT vs Logic-PPT & 0.904 & 0.000022 & 13 \\
\bottomrule
\end{tabular}
\caption{Pairwise Spearman correlations between task-emergence orders under different PPT conditions.}
\label{tab:spearman}
\end{table}

\clearpage

  \begin{table*}[t]
  \centering
  \scriptsize
  \setlength{\tabcolsep}{4pt}
  \renewcommand{\arraystretch}{1.15}
  \begin{tabularx}{\textwidth}{p{0.17\textwidth}p{0.28\textwidth}r>{\raggedright\arraybackslash}X}
  \toprule
  \textbf{Category} & \textbf{Task} & \textbf{\# Samples} & \textbf{Example} \\
  \midrule

  String Operations (72)
  & \texttt{copying} & 20
  & \textbf{Input:} gTpigTHK $\rightarrow$ \textbf{Output:} gTpigTHK \\
  & \texttt{simple\_icl:lowercase} & 26
  & \textbf{Input:} B $\rightarrow$ \textbf{Output:} b \\
  & \texttt{simple\_icl:uppercase} & 26
  & \textbf{Input:} b $\rightarrow$ \textbf{Output:} B \\
  \midrule

  Morphology (344)
  & \texttt{simple\_icl:present\_to\_gerund} & 179
  & \textbf{Input:} run $\rightarrow$ \textbf{Output:} running \\
  & \texttt{simple\_icl:singular\_to\_plural} & 165
  & \textbf{Input:} child $\rightarrow$ \textbf{Output:} children \\
  \midrule

  Reading Comprehension (1081)
  & \texttt{coreference:pronoun\_hard} & 20
  & \textbf{Input:} ``The trophy didn't fit in the suitcase because it was too big.'' What was too big? $\rightarrow$ \textbf{Output:} the trophy \\
  & \texttt{coreference:pronoun\_simple} & 20
  & \textbf{Input:} ``Alice told Bob that she would be late.'' Who does ``she'' refer to? $\rightarrow$ \textbf{Output:} Alice \\
  & \texttt{fact\_extraction:extract\_entity} & 20
  & \textbf{Input:} Alice gave five apples to Bob at the park. Who received the apples? $\rightarrow$ \textbf{Output:} Bob \\
  & \texttt{fact\_extraction:extract\_number} & 20
  & \textbf{Input:} John gave 5 apples to Mary on Tuesday. How many apples? $\rightarrow$ \textbf{Output:} 5 \\
  & \texttt{ignoring\_context} & 1
  & \textbf{Input:} Some text here. $X=5$. More text. What is $X$? $\rightarrow$ \textbf{Output:} 5 \\
  & \texttt{ioi\_task} & 1000
  & \textbf{Input:} Henry and Phil had a lot of fun at the harbor. Henry gave a basket to $\rightarrow$ \textbf{Output:} Phil \\
  \midrule

  Compositional (3090)
  & \texttt{blended\_compositions:extract\_verify} & 6
  & \textbf{Input:} Nora gave 3 apples to Ben; Ben gave 1 apple to Li. Claim: Ben received apples before giving any away. $\rightarrow$ \textbf{Output:} True \\
  & \texttt{compositional:extract\_verify} & 6
  & \textbf{Input:} Ravi arrived after Mina, but before Joel. Claim: Joel arrived before Mina. $\rightarrow$ \textbf{Output:} False \\
  & \texttt{compositional:lower\_first} & 971
  & \textbf{Input:} AFGHANISTAN $\rightarrow$ \textbf{Output:} a \\
  & \texttt{compositional:plural\_first} & 165
  & \textbf{Input:} child $\rightarrow$ \textbf{Output:} c \\
  & \texttt{compositional:reverse\_last} & 971
  & \textbf{Input:} Afghanistan $\rightarrow$ \textbf{Output:} A \\
  & \texttt{compositional:upper\_first} & 971
  & \textbf{Input:} afghanistan $\rightarrow$ \textbf{Output:} A \\

  \bottomrule
  \end{tabularx}
  \caption{Elemental tasks grouped by skill category, with the number of evaluation examples and one representative example for each task.}
  \label{tab:emergent-task-categories}
  \end{table*}

\end{document}